\documentclass{article}

\usepackage{microtype}
\usepackage{graphicx}
\usepackage{subcaption}
\usepackage{booktabs} 

\usepackage{hyperref}

\usepackage{float}

\usepackage[preprint]{icml2026}

\usepackage{amsmath}
\usepackage{amssymb}
\usepackage{mathtools}
\usepackage{amsthm}

\usepackage[capitalize,noabbrev]{cleveref}

\theoremstyle{plain}
\newtheorem{theorem}{Theorem}[section]
\newtheorem{proposition}[theorem]{Proposition}
\newtheorem{lemma}[theorem]{Lemma}
\newtheorem{corollary}[theorem]{Corollary}
\theoremstyle{definition}
\newtheorem{definition}[theorem]{Definition}
\newtheorem{assumption}[theorem]{Assumption}
\theoremstyle{remark}
\newtheorem{remark}[theorem]{Remark}

\usepackage[textsize=tiny]{todonotes}

\usepackage{natbib} 

\usepackage[acronym,nowarn]{glossaries}
\glsdisablehyper
\makeglossaries

\usepackage{xspace}

\usepackage{tikz}
\usetikzlibrary{positioning, arrows.meta, calc, fit, backgrounds}

\newacronym{FID}{\textsc{fid}}{Fr\'echet Inception Distance}

\newacronym{KL}{\textsc{kl}}{Kullback-Leibler}
\newcommand{\kl}{\gls{KL}\xspace}

\newacronym{MLP}{\textsc{mlp}}{Multi Layer Perceptron}
\newacronym{VAE}{\textsc{vae}}{Variational Autoencoder}
\newacronym{GAN}{\textsc{gan}}{Generative Adversarial Network}
\newacronym{WGAN}{\textsc{w-gan}}{Wasserstein Generative Adversarial Network}
\newacronym{CycleGAN}{\textsc{cycle-gan}}{Cycle Generative Adversarial Network}
\newacronym{MMDGAN}{\textsc{mmd-gan}}{Maximum Mean Discrepancy Generative Adversarial Network}
\newacronym{FGAN}{$f$-\textsc{gan}}{Generative Adversarial Networks with $f$-divergence}
\newacronym{DCGAN}{\textsc{dc-gan}}{Deep Convolutional Generative Adversarial Network}
\newacronym{RGAN}{\textsc{r-gan}}{Relativistic Generative Adversarial Network}
\newcommand{\vae}{\gls{VAE}\xspace}

\newcommand{\gan}{\gls{GAN}\xspace}
\newcommand{\gans}{\glspl{GAN}\xspace}
\newcommand{\wgan}{\gls{WGAN}\xspace}
\newcommand{\wgans}{\glspl{WGAN}\xspace}
\newcommand{\mmdgan}{\gls{MMDGAN}\xspace}
\newcommand{\mmdgans}{\glspl{MMDGAN}\xspace}

\newcommand{\fgans}{\glspl{FGAN}\xspace}
\newcommand{\dcgan}{\gls{DCGAN}\xspace}

\newcommand{\rgan}{\gls{RGAN}\xspace}
\newcommand{\rgans}{\glspl{RGAN}\xspace}
\newcommand{\cyclegans}{\textsc{cycle-gan}s\xspace}
\newcommand{\stylegantwo}{\textsc{style-gan2}\xspace}
\newcommand{\stylegantwoada}{\textsc{style-gan2-ada}\xspace}
\newcommand{\stylegan}{\textsc{style-gan}\xspace}
\newcommand{\rthreegan}{\textsc{r3-gan}\xspace}

\newacronym{INR}{\textsc{inr}}{Implicit Neural Representations}

\newcommand{\mlp}{\gls{MLP}\xspace}
\newcommand{\mlps}{\glspl{MLP}\xspace}

\newacronym{GMMN}{\textsc{gmmn}}{Generative Moment Matching Network}

\newacronym{MMD}{\textsc{mmd}}{Maximum Mean Discrepancy}
\newcommand{\mmd}{\gls{MMD}\xspace}

\newacronym[firstplural=Bayesian neural networks]{BNN}{\textsc{bnn}}{Bayesian neural network}
\newcommand{\bnn}{\gls{BNN}\xspace}
\newcommand{\bnns}{\glspl{BNN}\xspace}

\newacronym{DARTS}{\textsc{darts}}{Differentiable Architecture Search}
\newcommand{\darts}{\gls{DARTS}\xspace}

\newacronym{SAM}{\textsc{sam}}{Sharpness-Aware Minimization}
\newcommand{\sam}{\gls{SAM}\xspace}

\newacronym{EMA}{\textsc{ema}}{Exponential Moving Average}
\newcommand{\ema}{\gls{EMA}\xspace}

\newcommand{\iscscore}{\textsc{isc}\xspace}
\newcommand{\fidscore}{\textsc{fid}\xspace}
\newcommand{\kidscore}{\textsc{kid}\xspace}

\newacronym{PTC}{\textsc{ptc}}{Post-Training Compression}

\newacronym{MCD}{\textsc{mcd}}{Monte Carlo Dropout}
\newcommand{\mcd}{\gls{MCD}\xspace}

\newcommand{\cifar}{\textsc{cifar}-\textsc{{10}}\xspace}

\newcommand{\mnist}{\textsc{mnist}\xspace}
\newcommand{\celeba}{\textsc{celeba}\xspace}

\newcommand{\ffhqlowres}{\textsc{ffhq-128}\xspace}
\newcommand{\ffhqlowerres}{\textsc{ffhq256}\xspace}
\newcommand{\ffhq}{\textsc{ffhq}\xspace}

\newcommand{\mathbold}[1]{{\boldsymbol{\mathbf{#1}}}}

\renewcommand{\d}[1]{\ensuremath{\operatorname{d}\!{#1}}}
\newcommand{\nestedmathbold}[1]{{\mathbold{#1}}}

\newcommand{\mbf}{\nestedmathbold{f}}

\newcommand{\mbh}{\nestedmathbold{h}}

\newcommand{\mbu}{\nestedmathbold{u}}

\newcommand{\mbx}{\nestedmathbold{x}}

\newcommand{\mbz}{\nestedmathbold{z}}

\newcommand{\mbI}{\nestedmathbold{I}}

\newcommand{\mbX}{\nestedmathbold{X}}

\newcommand{\mbZ}{\nestedmathbold{Z}}

\newcommand{\mbepsilon}{\nestedmathbold{\epsilon}}

\newcommand{\mbeta}{\nestedmathbold{\eta}}

\newcommand{\mbpsi}{\nestedmathbold{\psi}}

\newcommand{\mbzero}{\nestedmathbold{0}}

\DeclareRobustCommand{\klmath}[2]{\ensuremath{\textsc{kl}\left[#1\;\|\;#2\right]}}
\DeclarePairedDelimiterX{\infdivx}[2]{[}{]}{%
  #1\;\delimsize\|\;#2%
}

\newcommand{\cD}{\mathcal{D}}
\newcommand{\cL}{\mathcal{L}}
\newcommand{\cN}{\mathcal{N}}

\newcommand{\cH}{\mathcal{H}}

\newcommand{\E}{\mathbb{E}}

\newcommand{\bbR}{\mathbb{R}}

\icmltitlerunning{Bridging GANs and Bayesian Neural Networks via Partial Stochasticity}

\begin{document}

\twocolumn[
  \icmltitle{Bridging GANs and Bayesian Neural Networks via Partial Stochasticity}
  


  \icmlsetsymbol{equal}{*}

  \begin{icmlauthorlist}
    \icmlauthor{Maurizio Filippone}{kaust}
    \icmlauthor{Marius P. Linhard}{rptu}
  \end{icmlauthorlist}

  \icmlaffiliation{kaust}{Statistics Program, KAUST, Saudi Arabia}
  \icmlaffiliation{rptu}{RPTU Kaiserslautern-Landau, Germany}
 
  \icmlcorrespondingauthor{Maurizio Filippone}{maurizio.filippone@kaust.edu.sa}

  \icmlkeywords{Generative Adversarial Networks, Bayesian Neural Networks, Bayesian Model Selection}

  \vskip 0.3in
]



\printAffiliationsAndNotice{}  

\begin{abstract}
\gans are popular and successful generative models.
Despite their success, optimization is notoriously challenging. 
In this work, we explain the success and limitations of \gans by casting them as Bayesian neural networks with partial stochasticity. 
This interpretation allows us to establish conditions of universal approximation and to rewrite the adversarial-style optimization of several variants of \gans as the optimization of a proxy for the likelihood obtained by marginalizing out the stochastic variables. 
Following this interpretation, the need for regularization becomes apparent, and we propose to adopt strategies to smooth the loss landscape and methods to search for solutions with minimum description length, which are associated with flat minima and good generalization.
Results obtained on a wide range of experiments indicate that these strategies lead to performance improvements and pave the way to a deeper understanding of \gans.
\end{abstract}

\section{Introduction}
\label{sec:introduction}

\glsresetall

\gans \citep{Goodfellow14} are a popular and powerful class of generative models originally conceived for artificial curiosity \citep{Schmidhuber90,Schmidhuber91}.
\gans have shown impressive performance, e.g., in generating realistic-looking images in computer vision applications (see, e.g., \citet{9156570, Wang23}).
A notoriously difficult aspect of \gans is their optimization, and we speculate that this is the reason why the literature on generative modeling has recently shifted its focus to diffusion models.
However, \gans remain attractive because once trained, the cost of generating one sample is as low as one model evaluation, while diffusion models require more computational effort \citep{Zheng23}. 

In this work, our aim is to revive the interest in \gans by studying them as \bnns with partial stochasticity \citep{Sharma23}, allowing us to advance our understanding of 
their success and limitations.
In our analysis, 
we introduce a set of random variables $p(\mbz) = \cN(\mbz | \mbzero, \mbI)$ that is transformed into a set of random variables $\mbx$ through a function $\mbf_{\mathrm{gen}}(\mbz, \mbpsi)$ parameterized by a neural network ({\em a.k.a.} the generator) with parameters $\mbpsi$ \citep{Bishop98,MacKay95}.
This construction is considered partially stochastic because all stochasticity is encapsulated in $\mbz$, while neural network parameters are deterministic (\emph{Model~b} in \cref{fig:graphical_models}). 

\begin{figure}[t]

\begin{tikzpicture}[
    node distance=0.55cm and 0.55cm,
    mynode/.style={draw, circle, minimum size=0.7cm, font=\footnotesize, align=center},
    obsnode/.style={mynode, fill=blue!30}, 
    dot/.style={fill, circle, inner sep=0pt, minimum size=1.5mm},
    arrow/.style={-Stealth, thick},
    plate/.style={draw, rectangle, inner sep=10pt, rounded corners=2pt, thick},
    labeltext/.style={below=0.3cm, font=\itshape, align=center}
]

    \begin{scope}[xshift=0.5cm, local bounding box=model1]
        \node[obsnode] (a1) {$\mathbf{x}$}; 
        \node[mynode, above=of a1] (b1) {$\mbpsi$};
        \node[dot, above=of b1, label=above:$\mathbf{h}$] (c1) {};
        
        \draw[arrow] (b1) -- (a1);
        \draw[arrow] (c1) -- (b1);
        
        \node[labeltext] at (a1.south) {Model a};

        \begin{scope}[on background layer]
            \node[plate, fit=(a1), label={[anchor=south east]south east:$N$}] (plate1) {};
        \end{scope}
    \end{scope}

    \begin{scope}[xshift=3.3cm, local bounding box=model2]
        \node[obsnode] (b2) {$\mathbf{x}$};
        \node[mynode, above=of b2] (a2) {$\mathbf{z}$};
        \node[dot, above right=1.85cm and 1.0cm of b2, label=above:$\mathbf{\mbpsi}$] (c2) {};
        
        \draw[arrow] (a2) -- (b2);
        \draw[arrow] (c2) -- (b2);
        
        \node[labeltext] at (b2.south) {Model b};

        \begin{scope}[on background layer]
            \node[plate, fit=(a2) (b2), label={[anchor=south east]south east:$N$}] (plate1) {};
        \end{scope}
    \end{scope}

    \begin{scope}[xshift=6.3cm, local bounding box=model3]
        \node[obsnode] (b3) {$\mathbf{x}$};
        \node[mynode, above=of b3] (a3) {$\mathbf{z}$};
        \node[mynode, above right=0.75cm and 0.75cm of b3] (c3) {$\mbpsi$};
        \node[dot, above=of c3, label=above:$\mathbf{h}$] (d3) {};
        
        \draw[arrow] (a3) -- (b3);
        \draw[arrow] (c3) -- (b3);
        \draw[arrow] (d3) -- (c3);
        
        \node[labeltext] at (b3.south) {Model c};
        
        \begin{scope}[on background layer]
            \node[plate, fit=(a3) (b3), label={[anchor=south east]south east:$N$}] (plate1) {};
        \end{scope}
    \end{scope}

\end{tikzpicture}

\caption{Graphical model representation of various \bnns studied in this work. 
We use the standard convention that nodes denote stochastic random variables, while dots represent deterministic ones. 
Shaded nodes denote observed variables, and we use the plate notation to indicate that certain random variables have $N$ repetitions. 
\emph{Model a} represents a \bnn with full stochasticity for generative modeling (e.g., \citet{TranNeurIPS21}). 
\emph{Model b} represents \bnns with partial stochasticity, where all stochasticity is captured by $\mbz$ \citep{Sharma23}; this is the construction that we use to study of \gans in this work. 
\emph{Model c} is the graphical model of Bayesian \gans \citep{Saatchi17}. 
}
\label{fig:graphical_models}

\end{figure}

The first insight into the success of \gans comes from the literature on \bnns \citep{Neal96,Mackay94}; we can adapt recent results on \bnns with partial stochasticity \citep{Sharma23} to establish that \gans are universal approximators of any absolutely continuous distribution over $\mbx$, provided that the dimensionality of latent variables is large enough and that the generator satisfies the standard conditions of universal function approximation \citep{Leshno93}.
In other words, partial stochasticity leads to sufficiently expressive \bnns, compared to fully stochastic \bnns where neural network parameters are considered stochastic and they have a prior conditioned on some hyper-parameters $\mbh$ \citep{TranNeurIPS21} (\emph{Model~a} in \cref{fig:graphical_models}).

The second insight into the success of \gans comes from the analysis of the likelihood of the corresponding graphical model. 
After defining the latent variables $\mbZ = \{ \mbz_i \}_{i=1, \ldots, N}$ associated with the training data $\mbX$, we would like to integrate these variables out as $\int p(\mbX | \mbZ, \mbpsi) p(\mbZ) d\mbZ$ to obtain the marginalized likelihood $p(\mbX | \mbpsi)$. 
However, the intractability of this objective prevents us from being able to optimize it with respect to $\mbpsi$.

In this paper, we show that this marginalized likelihood can be rewritten as the \kl divergence between $\pi(\mbx)$, the true generating distribution, and $p(\mbx | \mbpsi)$.
We can then derive the objective of many popular \gans by replacing the \kl divergence with alternative matching objectives. 
Interestingly, computing many popular matching objectives requires the definition and optimization of a discriminator, which then becomes an accessory to \gans optimization. 
Another notable aspect is that the objective is designed so that it can be optimized using samples from $\pi(\mbx)$, that is, our training set $\mbX$, and samples from the model $p(\mbx | \mbpsi)$, which are easy to obtain.
We can cast several \gans within our unified framework, and we report in particular on standard \gans \citep{Goodfellow14}, \fgans \citep{Nowozin16}, \wgans \citep{Arjovsky17}, and \mmdgans \citep{Dziugaite15,Li17}.


Our analysis of \gans as \bnns with partial stochasticity reveals their limitations too. 
\gans effectively perform some form of maximum likelihood. 
Viewed through these lenses, it is clear that there is no intrinsic mechanism to control overfitting, and if the model is too flexible it can assign large values of $p(\mbX | \mbpsi)$ on the training data, which is undesirable.

This realization motivates us to propose various ways to improve \gans. 
One line of investigation is to introduce regularization as a means to improve generalization, and we propose likelihood relaxation and gradient regularization \citep{Arjovsky17b,Roth17,Nagarajan17}. 
Another line of attack is to encourage optimization to obtain solutions associated with good generalization. 
The connection between Occam's razor, minimum description length (low Kolmogorov complexity), and flat minima \citep{Solomonoff64,Hochreiter97,Schmidhuber95}, motivates us to study ways to optimize parameters by guiding the optimization toward flat minima \citep{Hochreiter97,Foret21}. 
A third line of study is to treat model parameters in an approximate Bayesian way to reap their intrinsic benefits of regularization through ensembling \citep{Saatchi17} (\emph{Model C} in \cref{fig:graphical_models}).
In order to keep a practical implementation of \gans, we will adopt \mcd \citep{Gal16}, which has been shown to be an instance of variational inference and it is straightforward to adopt. 

Through extensive experiments, we support the conclusion that these strategies are effective in improving generative modeling performance in \gans. 
The main contributions of this paper are as follows:

{\bf Deriving \gans objectives from first principles.} In \cref{sec:gans_as_lvms}, we show that various popular \gans target tractable sample-based proxies for the intractable likelihood obtained by marginalizing out the random variables $\mbz$. 
More specifically, these proxies can be estimated through samples from the model and data;

{\bf Deriving approximation guarantees for \gans.} In \cref{sec:gans_as_bnns}, we build on recent results on \bnns with partial stochasticity to establish the conditions enabling \gans to be universal approximators of any absolutely continuous distribution over $\mbx$;

{\bf Understanding and improving \gans.} In \cref{sec:experiments}, we leverage our improved understanding of \gans to empirically demonstrate that model regularization, flat minima search, and approximate inference generally enable \gans to achieve higher generation quality compared to standard optimization. 

\section{Background}
\label{sec:background}

We consider a generative modeling task for a random variable $\mbx$ taking values in $\mathcal{X} \subseteq \mathbb{R}^D$, starting from a dataset $\mbX = \{ \mbx_1, \dots, \mbx_N \}$, where the samples $\mbx_i$ are drawn i.i.d. from an unknown distribution with associated density $\pi(\mbx)$.
Throughout this paper, we consider random variables whose probability measure is absolutely continuous with respect to the $D$-dimensional Lebesgue measure. To simplify the exposition, we use $\pi(\mbx)$ to denote both the probability distribution and its density function. Similarly, we use $\mbx$ to represent both the random variable and its realization.



\paragraph{Probabilistic Generative Models using Neural Networks}
We can set up a probabilistic model for this task as follows.
Let's introduce a set of latent variables $\mbZ = \{ \mbz_1, \ldots, \mbz_N \}$, with $\mbz_i \in \bbR^{P}$ and a parametric model $p(\mbx | \mbz, \mbpsi)$.
The parameters $\mbpsi$ refer to the ones of a neural network $\mbf_{\mathrm{gen}}(\mbz_i, \mbpsi)$ mapping latent variables $\mbz_i$ into corresponding $\mbx_i$. 
In a parallel with \gans, we consider a deterministic generator, so the likelihood can be written as:
\begin{equation}
p(\mbx_i | \mbz_i, \mbpsi) = \delta(\mbx_i - \mbf_{\mathrm{gen}}(\mbz_i, \mbpsi)) \text{,} 
\end{equation}
where $\delta$ is Dirac's delta.
 This construction takes the input distribution over latent variables $\mbz$ and turns it into a flexible distribution over $\mbx$.
 This class of latent variable models is known under several names in the literature \citep{MacKay95,Bishop98,Nowozin16}, and we will refer to these as generative neural samplers, or, more simply, as generators.
 In this work, we interpret these latent variable models as \bnns with partial stochasticity (\emph{Model b} in \cref{fig:graphical_models}). 
 In this interpretation, the random variables $\mbz$ in input are transformed into $\mbx$ through a mapping $\mbf_{\mathrm{gen}}(\mbz_i, \mbpsi)$; parameters are considered deterministic (to be optimized) and all the stochasticity of this model is captured by $\mbz$. 

Following common practice in latent variables models, we can attempt to integrate out the latent variables $\mbZ$
\begin{equation}
p(\mbX | \mbpsi) = \int p(\mbX | \mbZ, \mbpsi) p(\mbZ) \d\mbZ  \text{,} 
\end{equation}
with the aim of maximizing this objective with respect to $\mbpsi$.
A closer inspection of the integral above, under the factorization of the likelihood $p(\mbX | \mbZ, \mbpsi) = \prod_i p(\mbx_i | \mbz_i, \mbpsi)$ indicates that we can gracefully factorize the marginalized likelihood $p(\mbX | \mbpsi)$ as the product of individual data-specific marginalized likelihood terms: 
\begin{equation}
p(\mbX | \mbpsi) = \prod_i \int  p(\mbx_i | \mbz_i, \mbpsi) p(\mbz_i) \d\mbz_i =  \prod_i p(\mbx_i | \mbpsi)  \text{.} \nonumber
\end{equation}
Optimizing (the logarithm of) this objective directly is challenging because it is intractable to marginalize out latent variables in interesting scenarios where $D$ and $P$ are even moderately large and $\mbf_{\mathrm{gen}}(\mbz_i, \mbpsi)$ is implemented by a neural network. 
%
%
Up to a scaling factor of $1/N$, we can interpret this objective as a Monte Carlo average of the following expectation ({\em a.k.a.} expected risk):
\begin{equation} \label{eq:marginal_likelihood}
\mbpsi_* = \arg\max_{\mbpsi} \E_{\pi(\mbx)} \left\{ \log \left[ p(\mbx | \mbpsi) \right] \right\} \text{,}
\end{equation}
which we will use as a starting point of our analysis in the next section.

\section{A Practical Proxy for the Likelihood}
\label{sec:gans_as_lvms}

For reasons that will be apparent soon, let $P\geq D$.
Simple manipulations of \cref{eq:marginal_likelihood} show that the optimization of the marginalized likelihood can be rewritten equivalently as \citep{Akaike73,TranNeurIPS21}:
\begin{equation}
  \label{eq:marginal_likelihood_equivalence}
\mbpsi_*  = \arg\min_{\mbpsi}  \klmath{\pi(\mbx)}{p(\mbx | \mbpsi)} \text{.}
\end{equation}
The assumption that $P \geq D$ ensures that the divergence is not degenerate.
This result says that the optimal model is the one which minimizes the \kl divergence between the true generating distribution and the one characterized by our generative model.
However, this reformulation does not immediately simplify the optimization problem.
This is because we can only access samples from $\pi(\mbx)$, and there is no closed form for the density $p(\mbx | \mbpsi)$; while it is straightforward to obtain samples from $p(\mbx | \mbpsi)$, sample-based estimators of the KL divergence through samples typically have large variance \citep{FlamShepherd17,TranJMLR22}.

For completeness, here is how to obtain samples from the two distributions of interest.
For $\pi(\mbx)$, we have samples $\mbx_i$, that is our data.
For $p(\mbx | \mbpsi)$, we can sample $\mbz$ from $\cN(\mbz | \mbzero, \mbI)$ and then $\mbx$ from $p(\mbx | \mbz, \mbpsi)$; this yields samples from the joint $p(\mbx, \mbz | \mbpsi)$, which is what we need to obtain samples from $p(\mbx | \mbpsi)$ if we disregard $\mbz$.

\subsection{Replacing the \kl divergence with other divergences or integral probability metrics.}
We can exploit the equivalence between marginal likelihood optimization and the matching of $\pi(\mbx)$ and $p(\mbx | \mbpsi)$ to derive tractable objectives, which turn out to be the objectives of popular \gans.
In particular, we can replace the \kl divergence with alternatives aimed at achieving the same objective of matching $p(\mbx | \mbpsi)$ to the true generating distribution $\pi(\mbx)$.
Here we have a number of choices, and we can draw from the literature on other divergences (e.g., $f$-divergences \citep{Nguyen07,Gneiting07}) or integral probability metrics \citep{muller1997integral} (e.g., $1$-Wasserstein distance \citep{Villani16} or \mmd \citep{Gretton06}). 
The result is a series of \gan formulations, which we discuss shortly.

Note that, \cref{eq:marginal_likelihood_equivalence} establishes an alternative formulation for the optimization of the marginalized likelihood as the optimization of a \kl divergence.
It would be interesting to use this equivalence in the opposite direction and derive the ``generalized marginalized likelihoods'' stemming from the use of other divergences or integral probability metrics.
We find this challenging due to the form of the matching objectives that in general entangle $p(\mbx | \mbpsi)$ and $\pi(\mbx)$ in a way that prevents expressing the left hand side as an expectation over $\pi(\mbx)$ of a function of $p(\mbx | \mbpsi)$, which is needed to interpret this as an expected risk.
While this might be possible for some particular matching objectives, we leave this investigation for future works. 

Note also that the discriminator, which does not appear in the formulation of latent variable models or \bnns with partial stochasticity, becomes an essential component of \gans, as it is generally needed to calculate matching objectives. 

\paragraph{\gans.}

Up to some constants, the objective of the original \gans in \citet{Goodfellow14} is:
\begin{equation}
\mbpsi_* = 
\arg\min_{\mbpsi} \, 
\textsc{js}[\pi(\mbx) || p(\mbx | \mbpsi)] \text{,}
\end{equation}
where $\mathrm{JS}$ is the Jensen-Shannon divergence:
\begin{align}
\textsc{js}[p(\mbx) || q(\mbx)] = 
\frac{1}{2} \klmath{p(\mbx)}{a(\mbx)} + 
\frac{1}{2} \klmath{q(\mbx)}{a(\mbx)} \nonumber \text{,}
\end{align}
with $a(\mbx) = \frac{1}{2}(p(\mbx) + q(\mbx))$.

\paragraph{\fgans.}

\citet{Nowozin16} present a more general class of \gans with objectives derived from $f$-divergences
$$
\cD_{f}(\pi(\mbx) \| p(\mbx | \mbpsi)) = \int p(\mbx | \mbpsi) \, f\left(\frac{\pi(\mbx)}{p(\mbx | \mbpsi)} \right) d\mbx \text{,}
$$
for which the Jensen-Shannon divergence is a special case.
Their work leverages variational methods to tractably estimate $f$-divergences between distributions through samples \citep{Nguyen07}.

\paragraph{\wgans.}

Within the family of integral probability metrics, we find the popular $1$-Wasserstein distance.
If we replace the \kl divergence in \cref{eq:marginal_likelihood_equivalence} with this metric, we can cast optimization as:
$$
\arg\min_{\mbpsi} \left\{W_1\left(\pi(\mbx), p(\mbx | \mbpsi) \right) \right\} \text{.}
$$
We can use a dual formulation of the $1$-Wasserstein distance to obtain the following objective:
$$
\arg\min_{\mbpsi} \left\{
\sup_{\mathrm{Lip}{(f)} \leq 1} \left(
\E_{\pi(\mbx)}[f(\mbx)] - \E_{p(\mbx | \mbpsi)} [f(\mbx)]
\right)
\right\}  \text{.}
$$
This approach is essentially the one presented in the \wgan paper \citep{Arjovsky17}.
The discriminator is modeled as a neural network, and the Lipschitz condition can be imposed either by adding a regularization term to the discriminator \citep{Gulrajani17} or by construction \citep{Ducotterd24}.

\paragraph{\mmdgan.}

\mmd \citep{Gretton06} is another member of the family of integral probability measures.
Replacing the \kl divergence in \cref{eq:marginal_likelihood_equivalence} with the \mmd, we obtain a similar objective to the \wgan, except that the discriminator is now a function in a Reproducing Kernel Hilbert Space (\textsc{rkhs}) denoted by $\cH$:
$$
\arg\min_{\mbpsi} \left\{
\sup_{f \in \cH, \|f\|_{\cH} \leq 1} \left(
\E_{\pi(\mbx)}[f(\mbx)] - \E_{p(\mbx | \mbpsi)} [f(\mbx)]
\right)
\right\} \text{.}
$$
In practice, it is convenient to square the \mmd distance so that the objective has a closed form, and it can be expressed through the evaluation of the kernel function $k(\cdot, \cdot)$ with samples from the two distributions as input. 
The objective lends itself to an unbiased estimate over mini-batches \citep{Gretton12}. 
The use of \mmd as a matching objective was proposed in \citet{Dziugaite15}, who also provide generalization bounds of the resulting \mmdgan. 
For discussions on the choice/optimization of a kernel function and $\sqrt{\cL_{\mmd^2}}$ as a loss, we refer the reader to \citet{Li15,Li17,Dziugaite15}.


\subsection{Understanding Overfitting in \gans}

\begin{figure}[t]
\centering
\includegraphics[width=\linewidth]{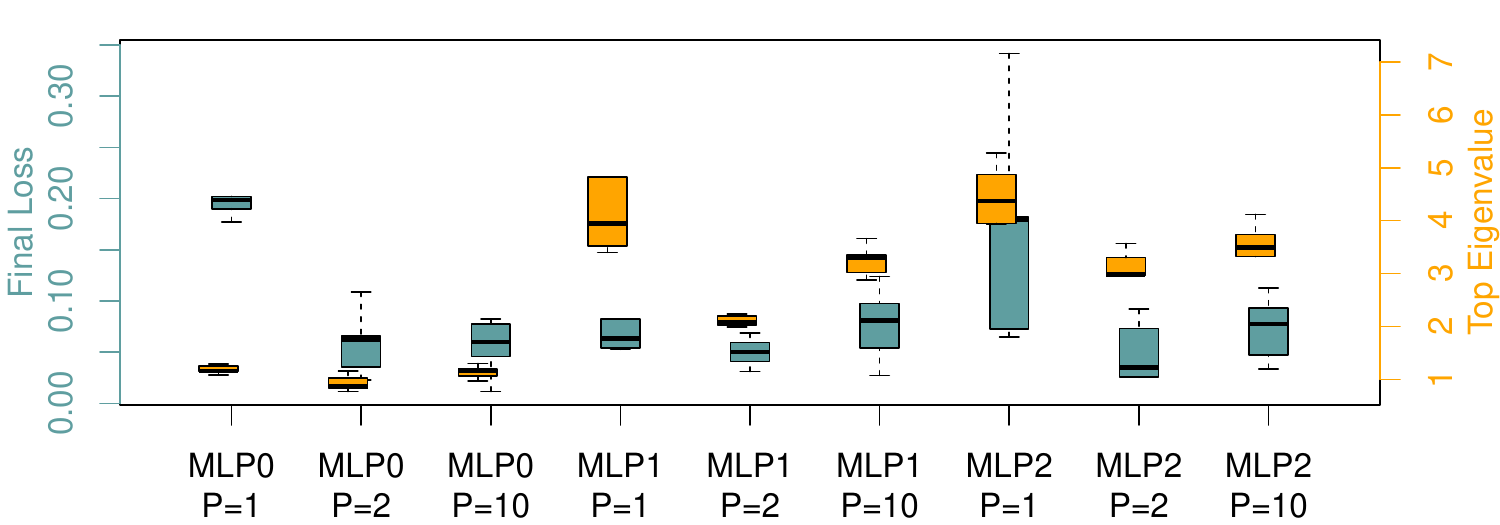}
\caption{\mmdgans on two-dimensional Gaussian data. 
Details on the experimental setup in the main text. }
\label{fig:2d_gaussian}
\end{figure}


We illustrate the issue of overfitting in \gans on a simple generative modeling problem, where the dataset contains $N = 2\,000$ input vectors drawn from a zero-mean and unit-variance Gaussian distribution with $D=2$.
We consider three possible latent dimensions $P = 1, 2, 10$ and three possible architectures for the generator: MLP0 indicates a \mlp with zero hidden layers (linear model), and MLP1 and MLP2 indicate \mlps with one and two hidden layers, respectively. 
The number of hidden units is set to $64$.
We train \mmdgans \citep{Dziugaite15}
with all possible combinations of latent dimensions and generator architectures. 
For the discriminator, we adopt an \mlp with two hidden layers with $64$ hidden units.
Each run is repeated five times. 

In \cref{fig:2d_gaussian}, we report a boxplot of the value of the objective obtained at the end of the optimization and a boxplot of the top eigenvalue of the Hessian of the objective. 
When the latent dimensionality is too low ($P<D$), the model is unable to attain good solutions, as indicated by the high value of the objective at the end of optimization, regardless of how complex the generator is. 
When $P>D$, all configurations reach a good solution, indicating a close match between the generated and true distributions. 
However, complex models are characterized by sharp minima (large top eigenvalue).
The model with the right level of complexity (MLP0 with $P=2$) shows that the solution obtained is indeed characterized by a flat loss landscape at the optimum.

\subsection{Regularization to Smooth Out the Loss Landscape}
\label{sec:model_regularization}

\paragraph{Likelihood relaxation}
One of the most striking features emerging from viewing \gans as latent variable models is that the likelihood is a degenerate Dirac's delta, with no aleatoric uncertainty.
Such a degeneracy of the likelihood is due to the constraint that one latent variable $\mbz$ has to be associated with one $\mbx$. 
A sensible relaxation is to turn the Dirac's delta into a Gaussian likelihood $\mathcal{N}(\mbx_i | \mbf_{\mathrm{gen}}(\mbz_i, \mbpsi), \sigma^2_{\mathrm{lik}})$, meaning that the generator produces $\mbx = \mbf_{\mathrm{gen}}(\mbz_i, \mbpsi) + \mbepsilon$ with $\mbepsilon \sim \mathcal{N}(\mbepsilon | 0, \sigma^2_{\mathrm{lik}})$ during training. 
Previous works have considered this form on noise perturbation to improve the stability of \gans optimization \citep{Arjovsky17b,Roth17}.
From a statistical perspective, we can understand this simple change as a likelihood relaxation aimed at improving model robustness, and we will demonstrate the effectiveness of this strategy in the experiments.
When generating images to evaluate performance, we just compute $\mbf_{\mathrm{gen}}(\mbz_i, \mbpsi)$ without adding noise.

\paragraph{Gradient Regularization}

Another technique to smooth the loss landscape in \gans is gradient regularization, which has been discussed, e.g., in \citet{Nagarajan17}:
$$
\hat{\mbpsi} = \arg\max_{\mbpsi} \left\{ \log \left[ p(\mbX | \mbpsi) \right] - \lambda_{\mathrm{gr}} \left\| \nabla_{\mbpsi} \log \left[ p(\mbX | \mbpsi) \right] \right\|^2 \right\} \text{.}
$$
For any \gans, the logarithm of the marginalized likelihood $\log \left[ p(\mbX | \mbpsi) \right]$ is then replaced by the corresponding generator lossc. 
The regularization pertains to the generator, so only its optimization is affected by this change.

\subsection{Searching for Flat Minima}
\label{sec:flat_minima_search}

\paragraph{Small batch sizes.}
One way to avoid sharp minima is to operate with noisy stochastic gradients, which can be achieved with small batch sizes. 
This strategy is a well-known implicit form of regularization (see, e.g., \citet{Brock19,Fatras20}). 
While a larger variance of the stochastic gradients adds to the instability of the optimization, various implementations of \gans available in the literature have indeed settled for a small batch size and a corresponding small learning rate, and we speculate that this is to reap the effects of the induced regularization while keeping the optimization stable to some extent.

\paragraph{Sharpness-Aware Minimization.}
\sam is a popular technique for searching for flat minima in the parameter space \citep{Foret21,Hochreiter97}. 
\sam operates by performing a standard stochastic gradient step, followed by a maximization of the objective in the neighborhood of radius $\rho_{\mathrm{SAM}}$. 
The rationale is that, in flat minima, the second step does not deteriorate the objective much, and the optimizer is encouraged to look for such solutions. 
We are not aware of previous attempts to use \sam in \gans.

\subsection{Approximate Inference}
\label{sec:mcd}

Another strategy to improve \gans is to carry out approximate inference of model parameters (\emph{Model c} in \cref{fig:graphical_models}). 
This was previously considered in \citet{Saatchi17}, where they adopt Markov chain Monte Carlo inference of the parameters of the generator and the discriminator. 
Given the huge computational cost associated with this approach, we consider \mcd as an alternative to carry out approximate (variational) inference of model parameters \citep{Gal16}. 
This can be easily implemented by introducing dropout layers in the generator and discriminator.

\section{Partially Stochastic Networks.}
\label{sec:gans_as_bnns}

We now establish approximation guarantees for \gans by casting them as \bnns with partial stochasticity. 
By adapting the theoretical framework in \citet{Sharma23}, we derive universal approximation properties for any absolutely continuous distribution $\pi(\mbx)$. 
To achieve this, we first define the requirements for the generator's architecture.


\begin{assumption} \label{assumption:uat}
The generator architecture $\mbf_{\mathrm{gen}}(\cdot, \mbpsi): \mathbb{R}^P \rightarrow \mathcal{X}$ satisfies the conditions in \citet{Leshno93}, ensuring it can approximate any continuous function $\tilde{\mbf}(\cdot): \mathbb{R}^P \rightarrow \mathcal{X}$ with arbitrary precision
\end{assumption}
Under this assumption, we present the main theorem establishing that \gans, viewed as \bnns with partial stochasticity, are universal approximators of probability distributions.

\begin{theorem} 
(Adapted from \citet{Sharma23}). \label{thm:uat_gan} 
Let $\mbx$ be a random variable in $\mathcal{X}\subseteq \mathbb{R}^D$ and
$\mbf_{\mathrm{gen}}(\cdot, \mbpsi): \mathbb{R}^P \rightarrow \mathcal{X}$ be a neural network satisfying \cref{assumption:uat}. 
Let $\mbz$ be Gaussian-distributed random variables with finite mean and variance.
If there exists a continuous generator function $\tilde{\mbf}(\cdot): \mathbb{R}^P \rightarrow \mathcal{X}$ that induces the distribution of $\mbx$ from the latent noise, then $\mbf_{\mathrm{gen}}(\cdot, \mbpsi)$ can approximate it arbitrarily well. 
Specifically, $\forall \varepsilon>0, \lambda < \infty$,
there exist parameters $\mbpsi \in \Psi$, a scaling matrix $V \in \mathbb{R}^{P\times P}$ and a shift-vector $\mbu \in \mathbb{R}^P$, such that
\begin{equation*} 
\sup_{\mbeta \in \mathbb{R}^P, |\mbeta| \le \lambda} |\mbf_{\mathrm{gen}}(V \mbeta + \mbu, \mbpsi) - \tilde{\mbf}(\mbeta)| < \varepsilon. 
\end{equation*} 
\end{theorem}

The proof, detailed in \citet{Sharma23}, leverages the noise outsourcing lemma \citep{Austin15} to guarantee the existence of $\tilde{\mbf}(\cdot)$, and the universal approximation theorem \citep{Leshno93} to guarantee that the network can represent it.


Informally, these conditions require enough stochasticity in $\mbz$ (e.g., $P$ large enough) so that the distribution produced by the generator can be mapped to the support of $\mbx$; in addition, the generator needs to have enough flexibility to be able to transform the distribution over $\mbz$ into any absolutely continuous distribution with density on the support of $\mbx$, and this is ensured by the classic universal approximation theorem for neural networks.
In practice, the manifold hypothesis \citep{LoaizaGanem22,Brown23} suggests that most large-dimensional datasets live in a low-dimensional manifold, which then relaxes the need to set $P \geq D$, and indeed in practice \gans work extremely well with $P \ll D$ for such applications.

\subsection{Graphical Models and Model Selection}

The theory helps us to establish conditions for ensuring universal approximation; however, the theory does not give practical advice on how to precisely determine the architecture and how to introduce the necessary stochasticity.
In \cref{fig:graphical_models}, under the right conditions, the three \bnn models have the same expressivity in approximating $\pi(\mbx)$. 
\emph{Model a} is the Bayesian AutoEncoder in \citet{TranNeurIPS21}, where the generator is simply an AutoEncoder whose parameters are inferred (through Markov chain Monte Carlo sampling) 
rather than optimized.
In this case, full stochasticity is encoded in the parameters, while in \gans, partial stochasticity is encapsulated in $\mbz$ (\emph{Model b} in \cref{fig:graphical_models}).
\emph{Model c} in \cref{fig:graphical_models} is the Bayesian \gan in \citet{Saatchi17}.
Again, the theory of \bnns with partial stochasticity does not favor one approach over the others, as long as the corresponding architectures satisfy \cref{assumption:uat} and $P$ is large enough, so model selection should ultimately be used to determine these choices.

\section{Related Works}
\label{sec:related_works}

 \paragraph{Improving the training dynamics of \gans.}
Training \gans is notoriously challenging, and divergent optimization dynamics is a common problem. 
\gans are supposed to converge to a Nash equilibrium, but this may not always exist \citep{pmlr-v119-farnia20a}. 
In order to alleviate this problem, there are various lines of work. 
\citet{pmlr-v119-farnia20a} propose to relax the constraint of Nash equilibrium and introduce a new training algorithm accordingly. 
Focusing on optimization but without relaxing the Nash equilibrium, \citet{pmlr-v115-nie20a} propose a way to regularize the Jacobian of the training dynamics.
\citet{NEURIPS2020_a851bd0d} propose the Top-k Training, where only the best generated samples are used to perform updates and train the generator. \citet{NEURIPS2024_4e2acb1e}, on the other hand, combine both architectural changes and gradient regularization of the discriminator to improve the training dynamics of \rgans \citep{jolicoeur-martineau2018}. Adapting the architecture in the \stylegantwo paper \citep{9156570} they introduce a new baseline for \gans, which they name \rthreegan. 
After proving that this indeed leads to local convergence, \citet{NEURIPS2024_4e2acb1e} empirically show that this new baseline is able to achieve state-of-the-art performance.

\paragraph{\gans as probabilistic generative models.} 

We view \gans as probabilistic generative models, where a set of latent variables is mapped to the input space through a neural network \citep{MacKay95,Bishop98}. 
In this interpretation, these are instances of \bnns \citep{Neal96,Mackay94} with partial stochasticity \citep{Sharma23}.
Unlike previous work on generative models with full network stochasticity \citep{Saatchi17,TranNeurIPS21}, in the probabilistic view of \gans, network parameters are treated deterministically and epistemic uncertainty is captured by the distribution over latent variables.
\citet{TranNeurIPS21} consider generative models in the form of auto-encoders, and more works exploring the connections between \gans and auto-encoders (\vae in particular) include \citet{Mescheder17,Balaji19}. 
It is worth mentioning previous work by \citet{Tiao18}, who carry out a variational analysis of latent variable models with an implicitly-defined prior over latent variables, which leads to a family of models that includes \cyclegans \citep{Zhu17} as a special case. 


\paragraph{Regularizing the generator.}
Regularization is a successful strategy to improve \gans optimization.
In the literature, however, a lot of effort has been dedicated to the improvement of statistical properties of the discriminator to improve optimization stability \citep{Gulrajani17,Wu18}.
Our work suggests that regularization plays an important role in improving the statistical properties of the generator. 
For instance, gradient norm regularization of the generator has been studied in \citet{Nagarajan17}, while adding noise to the generated samples has been considered in \citet{Arjovsky17,Roth17}.

\paragraph{Latest architectures and objectives.}
\citet{DBLP:journals/corr/abs-1812-04948} propose the \stylegan architecture, which forms the basis of state-of-the-art \gan models.
They consider the \rgan objective and propose a novel mechanism to handle the latent variables by introducing them within the layers of the generator. 
\stylegantwo \citep{9156570} was later proposed as an improvement over \stylegan, by tackling the problem of artifacts in the generated images through regularization and architectural improvements. 
\stylegantwo was then further improved in \citet{NEURIPS2020_8d30aa96} through an adaptive discriminator augmentation (\stylegantwoada), and in \citet{NEURIPS2021_076ccd93} using Fourier features, which improves generating quality for videos.

\section{Experiments}
\label{sec:experiments}

\subsection{Deep Convolutional \gans} 

\begin{table*}[t]
  \caption{\dcgan architecture with Wasserstein divergence objective \citep{Wu18} on \mnist, \cifar, \ffhqlowres, and \celeba. Standard deviations, calculated over five seeds times three repetitions of sampling $10\,000$ images, are reported in parenthesis.
  The arrows indicate whether metrics are so that the higher the better ($\uparrow$) or the lower the better ($\downarrow$).
  $|B|$ denotes the batch size. 
  }
  \label{tab:res:summary}
  \begin{center}
    \begin{small}
      \begin{sc}
\resizebox{\textwidth}{!}{\begin{tabular}{l|lll|lll}
    \toprule
    \multicolumn{7}{c}{\wgan \quad $|B|=128$} \\ 
    \midrule
     & \multicolumn{3}{c|}{\mnist} &  \multicolumn{3}{c}{\cifar} \\
    \midrule    
  Strategy  &   
    \iscscore $\uparrow$   &  \fidscore $\downarrow$  &  \kidscore {\scriptsize $\, \times 10^{-3} \,$} $\downarrow$  & 
    \iscscore $\uparrow$   &  \fidscore $\downarrow$  &  \kidscore {\scriptsize $\, \times 10^{-3} \,$} $\downarrow$  \\ 
    \midrule
None \quad {$\blacktriangle$}	 & 2.27 (0.02) 	 & 8.4 (0.4) 	 & 6.0 (0.4)   & 4.64 (0.13) 	 & 40.1 (1.5) 	 & 30.3 (1.6) 		 \\
Likelihood relax. ($\sigma^2_{\mathrm{lik}} = 0.01$)	 & 2.27 (0.02) 	 & 8.0 (0.4) 	 & 5.5 (0.4)    & 4.74 (0.08) 	 & 34.1 (2.2) 	 & 24.1 (2.1) 	 \\
Gradient reg.	($\lambda_{\mathrm{gr}} = 0.01$) & 2.26 (0.01) 	 & 8.2 (0.4) 	 & 5.8 (0.4)   & 4.72 (0.08) 	 & 37.8 (1.3) 	 & 27.9 (1.6) 	 \\
\sam ($\rho_{\mathrm{SAM}} = 0.1$)	 & 2.25 (0.02) 	 & 8.6 (0.4) 	 & 6.2 (0.5)    & 4.68 (0.15) 	 & 40.6 (3.4) 	 & 31.2 (3.5)  \\
\mcd ($p = 0.1$) \quad {$\triangle$}	 & 2.25 (0.02) 	 & 7.7 (0.5) 	 & 5.4 (0.5)    & 4.75 (0.05) 	 & 38.6 (2.2) 	 & 28.7 (2.2) 	 \\   
    \midrule
    \midrule
     & \multicolumn{3}{c|}{\ffhqlowres} &  \multicolumn{3}{c}{\celeba} \\
    \midrule    
Strategy  &   
    \iscscore $\uparrow$   &  \fidscore $\downarrow$  &  \kidscore {\scriptsize $\, \times 10^{-3} \,$} $\downarrow$  & 
    \iscscore $\uparrow$   &  \fidscore $\downarrow$  &  \kidscore {\scriptsize $\, \times 10^{-3} \,$} $\downarrow$  \\ 
    \midrule
None \quad {$\blacktriangle$}	 & 3.21 (0.05) 	 & 63.6 (1.5) 	 & 55.8 (1.5) 	  & 2.87 (0.04) 	 & 18.3 (0.9) 	 & 12.4 (1.1) 	 \\
Likelihood relax. ($\sigma^2_{\mathrm{lik}} = 0.01$) \quad {$\triangle$}	 & 3.25 (0.04) 	 & 62.8 (1.5) 	 & 54.8 (1.6)	    & 2.89 (0.05) 	 & 17.8 (2.6) 	 & 11.5 (2.7)  \\
Gradient reg.	($\lambda_{\mathrm{gr}} = 0.01$)	 & 3.22 (0.04) 	 & 65.1 (3.5) 	 & 57.0 (3.4)   & 2.88 (0.03) 	 & 18.5 (0.9) 	 & 12.6 (1.1)  \\
\sam ($\rho_{\mathrm{SAM}} = 0.1$)	 & 3.26 (0.05) 	 & 62.1 (2.0) 	 & 53.9 (2.3)	    & 2.84 (0.09) 	 & 34.9 (20.8) 	 & 28.5 (20.1) 	\\
\mcd ($p = 0.05$) & 3.20 (0.06) 	 & 65.5 (3.6) 	 & 57.4 (3.6)	     & 2.95 (0.13) 	 & 42.9 (19.4) 	 & 34.9 (18.2) 	 \\
   \bottomrule
  \end{tabular}}
  \end{sc}
  \end{small}
  \end{center}
\end{table*}

\begin{figure*}[t]
\begin{center}
\setlength{\tabcolsep}{3pt} 
\begin{tabular}{cccc}
{\bf Uncurated samples {$\blacktriangle$} }
& 
{\bf Uncurated samples {$\triangle$} }
&
{\bf Uncurated samples {$\blacktriangle$} }
& 
{\bf Uncurated samples {$\triangle$} }
\\
    \begin{tabular}{c}
      \includegraphics[scale=0.3]{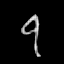}  
      \includegraphics[scale=0.3]{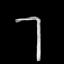}  
      \includegraphics[scale=0.3]{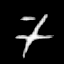}  
      \includegraphics[scale=0.3]{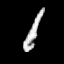}  
      \includegraphics[scale=0.3]{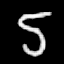}  \\
      \includegraphics[scale=0.3]{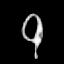}  
      \includegraphics[scale=0.3]{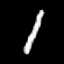}  
      \includegraphics[scale=0.3]{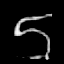}  
      \includegraphics[scale=0.3]{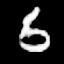}  
      \includegraphics[scale=0.3]{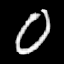}  \\
      \includegraphics[scale=0.3]{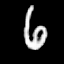}  
      \includegraphics[scale=0.3]{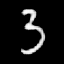}  
      \includegraphics[scale=0.3]{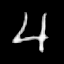}  
      \includegraphics[scale=0.3]{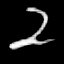}  
      \includegraphics[scale=0.3]{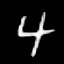}  \\
      \includegraphics[scale=0.3]{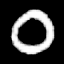}  
      \includegraphics[scale=0.3]{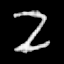}  
      \includegraphics[scale=0.3]{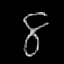}  
      \includegraphics[scale=0.3]{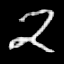}  
      \includegraphics[scale=0.3]{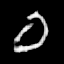}  \\
    \end{tabular}
      & 
    \begin{tabular}{c}
      \includegraphics[scale=0.3]{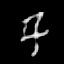}  
      \includegraphics[scale=0.3]{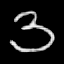}  
      \includegraphics[scale=0.3]{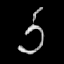}  
      \includegraphics[scale=0.3]{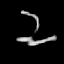}  
      \includegraphics[scale=0.3]{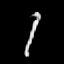}  \\
      \includegraphics[scale=0.3]{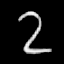}  
      \includegraphics[scale=0.3]{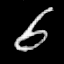}  
      \includegraphics[scale=0.3]{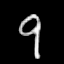}  
      \includegraphics[scale=0.3]{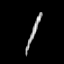}  
      \includegraphics[scale=0.3]{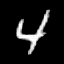}  \\
      \includegraphics[scale=0.3]{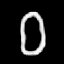}  
      \includegraphics[scale=0.3]{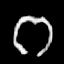}  
      \includegraphics[scale=0.3]{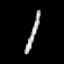}  
      \includegraphics[scale=0.3]{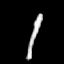}  
      \includegraphics[scale=0.3]{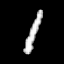}  \\
      \includegraphics[scale=0.3]{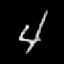}  
      \includegraphics[scale=0.3]{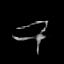}  
      \includegraphics[scale=0.3]{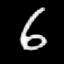}  
      \includegraphics[scale=0.3]{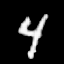}  
      \includegraphics[scale=0.3]{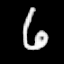}  \\
    \end{tabular}
&
    \begin{tabular}{c}
      \includegraphics[scale=0.15]{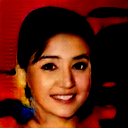}  
      \includegraphics[scale=0.15]{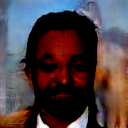}  
      \includegraphics[scale=0.15]{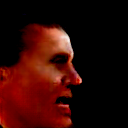}  
      \includegraphics[scale=0.15]{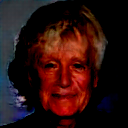}  
      \includegraphics[scale=0.15]{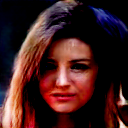}  \\
      \includegraphics[scale=0.15]{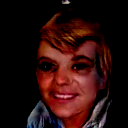}  
      \includegraphics[scale=0.15]{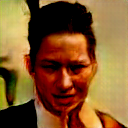}  
      \includegraphics[scale=0.15]{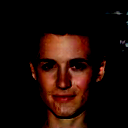}  
      \includegraphics[scale=0.15]{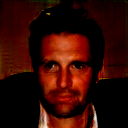}  
      \includegraphics[scale=0.15]{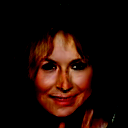}  \\
      \includegraphics[scale=0.15]{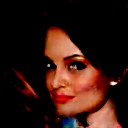}  
      \includegraphics[scale=0.15]{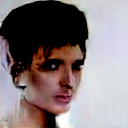}  
      \includegraphics[scale=0.15]{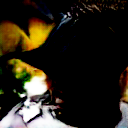}  
      \includegraphics[scale=0.15]{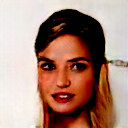}  
      \includegraphics[scale=0.15]{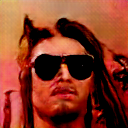}  \\
      \includegraphics[scale=0.15]{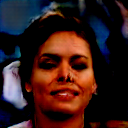}  
      \includegraphics[scale=0.15]{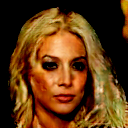}  
      \includegraphics[scale=0.15]{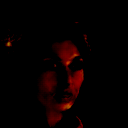}  
      \includegraphics[scale=0.15]{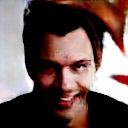}  
      \includegraphics[scale=0.15]{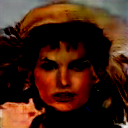}  \\
    \end{tabular}
      & 
    \begin{tabular}{c}
      \includegraphics[scale=0.15]{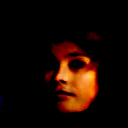}  
      \includegraphics[scale=0.15]{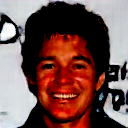}  
      \includegraphics[scale=0.15]{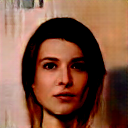}  
      \includegraphics[scale=0.15]{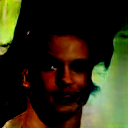}  
      \includegraphics[scale=0.15]{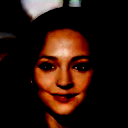}  \\
      \includegraphics[scale=0.15]{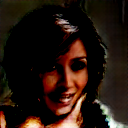}  
      \includegraphics[scale=0.15]{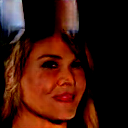}  
      \includegraphics[scale=0.15]{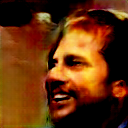}  
      \includegraphics[scale=0.15]{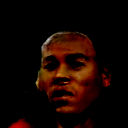}  
      \includegraphics[scale=0.15]{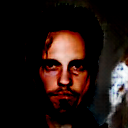}  \\
      \includegraphics[scale=0.15]{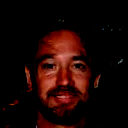}  
      \includegraphics[scale=0.15]{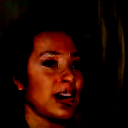}  
      \includegraphics[scale=0.15]{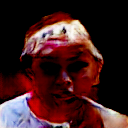}  
      \includegraphics[scale=0.15]{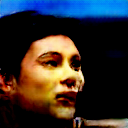}  
      \includegraphics[scale=0.15]{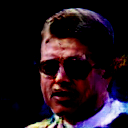}  \\
      \includegraphics[scale=0.15]{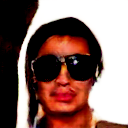}  
      \includegraphics[scale=0.15]{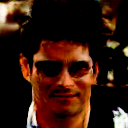}  
      \includegraphics[scale=0.15]{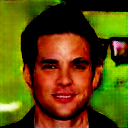}  
      \includegraphics[scale=0.15]{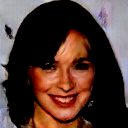}  
      \includegraphics[scale=0.15]{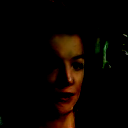}  \\
    \end{tabular}

\end{tabular}
\caption{\mnist and \celeba - uncurated samples generated from the models in \cref{tab:res:summary}.}
\label{fig:uncurated:mnist:celeba}
\end{center}
\end{figure*}

In this section, we validate the hypothesis that introducing regularization or searching for flat minima improves generative modeling performance.
We consider experiments on standard benchmark data, including \mnist, \cifar, \ffhq\footnote{\tiny  \url{https://github.com/NVlabs/ffhq-dataset.git}}, and \celeba\footnote{\tiny \url{https://mmlab.ie.cuhk.edu.hk/projects/CelebA.html}}. 
All models are optimized for $250$ epochs, except for \celeba for which we train models for $150$ epochs. 
We rescaled the images in \mnist and \cifar to $64 \times 64$, and the ones in \ffhq and \celeba to $128 \times 128$. 
Scores are computed over $10\,000$ generated images and averaged over three repetitions using the \texttt{torch-fidelity} module\footnote{\tiny \url{https://github.com/toshas/torch-fidelity}} against \texttt{.png} images resized as above. 
The results are reported by applying \ema, with $20$ epochs of warm-up and $0.999$ decay, reporting standard metrics, such as \iscscore, \fidscore, and \kidscore.
In the Appendix, we also include the top eigenvalue of the Hessian of the objective, estimated using the module \texttt{pytorch-hessian-eigenthings}\footnote{\tiny \url{https://github.com/noahgolmant/pytorch-hessian-eigenthings}}.
All experiments are repeated over five different seeds, and we report mean and standard deviation in the tables.

Throughout the experiments, we fix the architecture to be the one proposed in the \dcgan paper \citep{Radford15}. 
We report results on this architecture with the \wgan objective with divergence regularization \citep{Wu18} in \cref{tab:res:summary}, while in the Appendix we report results for \rgans, which use the so-called relativistic objective \citep{jolicoeur-martineau2018}. 
For completeness, in the Appendix we also provide a more comprehensive view of the results obtained by \wgans on a much broader set of parameters.

We fix the dimensionality of the latent space to $P = 100$.
For \wgans, we set the base learning rate to $0.001$ for a batch-size $|B|$ of $128$; 
the learning rate schedule is such that it reaches the base learning rate after $10$ epochs. 
For \rgans, we follow the recommendations in previous implementations and set the learning rate to $0.0002$.
In the experiments where $\rho_{\mathrm{SAM}} > 0$, we use (adaptive) \sam for both the generator and the discriminator.
In \wgans, we perform one optimization step for the generator every $5$ optimization steps for the discriminator, while for \rgans we do this after every optimization step of the discriminator. 
\mcd uses one Monte Carlo sample per training iteration and a dropout probability $p_{\mcd} = 0.1$ for \mnist and \cifar with dropout layers introduced after the second and third convolutional layers, while $p_{\mcd} = 0.05$ for \ffhqlowres and \celeba with dropout layers introduced after the second, third, and fourth convolutional layers. 
In the Appendix we report results with five Monte Carlo samples per training iteration. 
\cref{fig:uncurated:mnist:celeba} show samples associated with some of the configurations reported in \cref{tab:res:summary}.


For \mnist and \cifar, \cref{tab:res:summary} and \cref{tab:res:all_mnist_cifar} show that, unlike flat minima search, regularization offers consistent improvements. 
For \ffhqlowres and \celeba (\cref{tab:res:summary} and \cref{tab:res:all_ffhq128_celeba}), likelihood relaxation improves performance; gradient regularization seems to be effective for a smaller value of the regularization parameter ($\lambda_{\mathrm{gr}} = 0.001$).

The results using \sam optimization in \cref{tab:res:summary} do not show consistent improvement, indicating that there might be cases for which the simplicity associated with these solutions leads to suboptimal performance for \gans.
However, the full set of results in \cref{tab:res:all_mnist_cifar} and \cref{tab:res:all_ffhq128_celeba} also indicate that considerable improvements can be obtained by suitable regularization in combination with \sam. 
This is particularly the case for \cifar and \celeba. 

On \mnist and \cifar, \mcd obtains a striking consistent improvement in performance. 
For \ffhqlowres these improvements are less pronounced, while for \celeba \mcd negatively affects performance.

\subsection{\stylegantwoada}
\begin{figure}[t]
\begin{center}
\setlength{\tabcolsep}{3pt} 
\begin{tabular}{cc}
{\bf Uncurated samples {$\blacktriangle$} }
& 
{\bf Uncurated samples {$\triangle$} } \\
    \begin{tabular}{c}
          \includegraphics[scale=0.2]{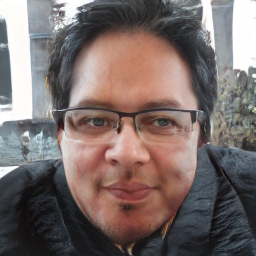}  
           \includegraphics[scale=0.2]{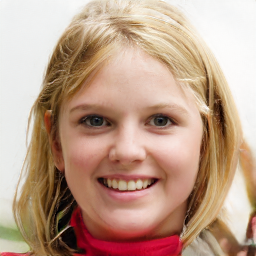}  \\  
           \includegraphics[scale=0.2]{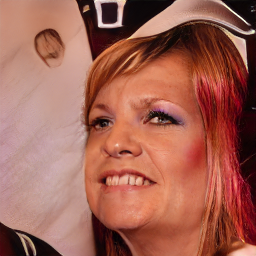}  
           \includegraphics[scale=0.2]{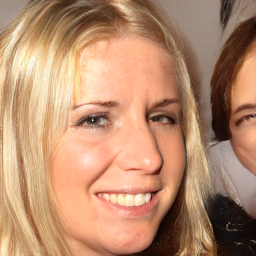}    \\
            \includegraphics[scale=0.2] {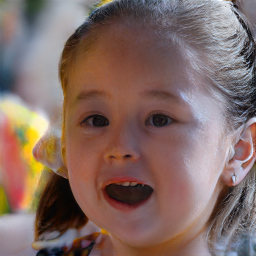}  
           \includegraphics[scale=0.2]{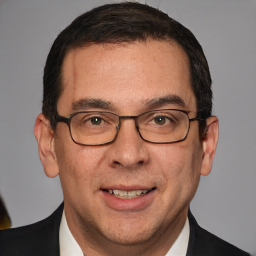}   
    \end{tabular}
      & 
    \begin{tabular}{c}
            \includegraphics[scale=0.2]{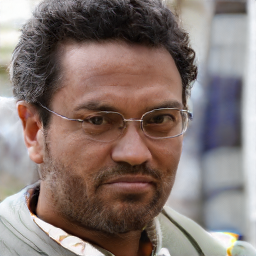}  
           \includegraphics[scale=0.2]{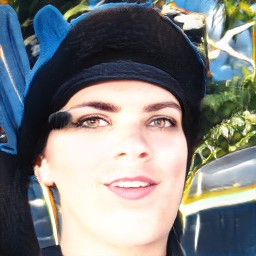}  \\
           \includegraphics[scale=0.2]{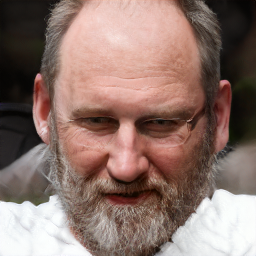}   
           \includegraphics[scale=0.2]{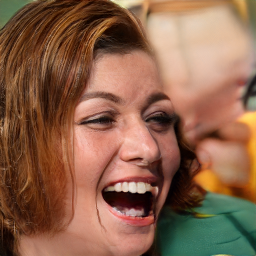}   \\
            \includegraphics[scale=0.2]{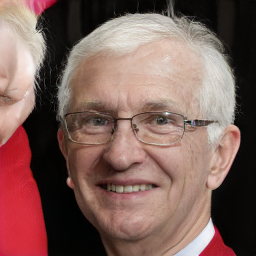} 
           \includegraphics[scale=0.2]{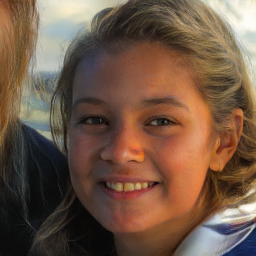}  
    \end{tabular}

\end{tabular}
\caption{Uncurated samples from the baseline and the one with likelihood relaxation on the \ffhqlowerres data.}
\label{fig:uncurated:sg2}
\end{center}
\end{figure}

In this section, we report experiments on \stylegantwoada on the \ffhq dataset rescaled to $256 \times 256$ (\ffhqlowerres). 
For this experiment, the baseline is \stylegantwoada with the same configuration as \citet{NEURIPS2020_8d30aa96}. 
We test the effect of likelihood relaxation by adding Gaussian noise $\mathcal{N}(0, \sigma_{\mathrm{lik}}^2)$ with $\sigma_{\mathrm{lik}}^2 = 0.001$ to the generated images during training. 
We train both models until the discriminator has seen $25$ million images, and the final \fidscore is 4.30 for the baseline model $(\blacktriangle)$ and 4.22 for the one with likelihood relaxation $(\triangle)$; samples from the two models can be found in \cref{fig:uncurated:sg2}. 
It is interesting to see how adding noise to the generating process indeed leads to performance improvements. 
This can also be observed after the models see another $5$ million images; plain \stylegantwoada manages to reduce the \fidscore to 4.11 while our model reaches an \fidscore of 4.07.
This experiment suggests that a simple modification to existing implementations of \gans can lead to performance improvements. 

\section{Conclusions}
\label{sec:conclusions}

In this paper, we proposed a probabilistic framework based on \bnns with partial stochasticity to understand and improve \gans. 
This allowed us
%
to establish universal approximation properties of \gans, and to derive the objective of a variety of popular \gans, where the intractable marginalized likelihood objective is replaced by a tractable proxy. 
This connection gives insights into overfitting, which manifests itself due to the lack of intrinsic regularization. 
By relying on the connections between Occam's razor, flat minima, and minimum description length, we studied regularization and optimization strategies to smooth the loss landscape of \gans and to search for flat minima.  
Overall, while there is no consistent pattern of which configuration is generally superior, the results indicate that the best performance is systematically achieved by configurations associated with model regularization, possibly in combination with \sam, and that in certain cases the performance improvement is rather substantial.
We hope that this will serve as a starting point for studying strategies to improve the statistical properties of \gans to obtain novel ways to boost their performance and ease optimization.
In this work, we kept the \gan architecture fixed, we are interested in studying architecture search in the future. 
For this, it would be interesting to explore approaches such as \darts \citep{Liu19}, possibly in combination with recent works introducing sparsity in \gans \citep{Wang23b}.



\paragraph{Limitations.}
Despite the encouraging results, 
it would be great to find ways to systematically obtain stable optimization and improved performance. 
For this, it would have been interesting to explore more \gan architectures, objectives, and hyper-parameters to derive general practical guidelines on how to guide these choices. 
Also, 
we were hoping to obtain stronger indications on the link between flat minima and good performance; despite this, performance is generally in favor of configurations associated with regularization, possibly combined with \sam optimization.  


\section*{Impact Statement}

This paper aims to advance the theoretical understanding and practice of \gans. 
Like other works in the literature of generative modeling, the proposed advancements can have both positive (e.g., synthesizing new data automatically, accelerating drug discovery) and negative (e.g., deep fakes) impacts on society depending on the application.

\section*{Acknowledgements}
MF is grateful to J\"urgen Schmidhuber for his wonderful seminar series at KAUST and for inspiring several ideas in this paper.

\bibliography{biblio.bib, filippone.bib, linhard.bib}

@article{Austin15,
     author = {Austin, Tim},
     title = {Exchangeable random measures},
     journal = {Annales de l'I.H.P. Probabilit\'es et statistiques},
     pages = {842--861},
     publisher = {Gauthier-Villars},
     volume = {51},
     number = {3},
     year = {2015},
     doi = {10.1214/13-AIHP584},
     mrnumber = {3365963},
     zbl = {1323.60068},
     language = {en},
     url = {https://www.numdam.org/articles/10.1214/13-AIHP584/}
}

@article{Leshno93,
title = {Multilayer feedforward networks with a nonpolynomial activation function can approximate any function},
journal = {Neural Networks},
volume = {6},
number = {6},
pages = {861-867},
year = {1993},
issn = {0893-6080},
doi = {https://doi.org/10.1016/S0893-6080(05)80131-5},
url = {https://www.sciencedirect.com/science/article/pii/S0893608005801315},
author = {Moshe Leshno and Vladimir Ya. Lin and Allan Pinkus and Shimon Schocken},
}

@article{Ducotterd24, author = {Ducotterd, Stanislas and Goujon, Alexis and Bohra, Pakshal and Perdios, Dimitris and Neumayer, Sebastian and Unser, Michael}, title = {{Improving Lipschitz-constrained neural networks by learning activation functions}}, year = {2024}, issue_date = {January 2024}, publisher = {JMLR.org}, volume = {25}, number = {1}, issn = {1532-4435}, journal = {J. Mach. Learn. Res.}, month = jan, articleno = {65}, numpages = {30} }

@inproceedings{Radford15,
  author       = {Alec Radford and
                  Luke Metz and
                  Soumith Chintala},
  title        = {{Unsupervised Representation Learning with Deep Convolutional Generative
                  Adversarial Networks}},
  booktitle    = {4th International Conference on Learning Representations},
  year         = {2016},
}

@inproceedings{
Brock19,
title={{Large Scale {GAN} Training for High Fidelity Natural Image Synthesis}},
author={Andrew Brock and Jeff Donahue and Karen Simonyan},
booktitle={International Conference on Learning Representations},
year={2019},
url={https://openreview.net/forum?id=B1xsqj09Fm},
}

@inproceedings{Wu18, author = {Wu, Jiqing and Huang, Zhiwu and Thoma, Janine and Acharya, Dinesh and Van Gool, Luc}, title = {{Wasserstein Divergence for GANs}}, year = {2018}, isbn = {978-3-030-01227-4}, publisher = {Springer-Verlag}, address = {Berlin, Heidelberg}, url = {https://doi.org/10.1007/978-3-030-01228-1_40}, doi = {10.1007/978-3-030-01228-1_40}, booktitle = {Computer Vision – ECCV 2018: 15th European Conference, Munich, Germany, September 8–14, 2018, Proceedings, Part V}, pages = {673–688}, numpages = {16}, location = {Munich, Germany} }

@inproceedings{Roth17,
 author = {Roth, Kevin and Lucchi, Aurelien and Nowozin, Sebastian and Hofmann, Thomas},
 booktitle = {Advances in Neural Information Processing Systems},
 editor = {I. Guyon and U. Von Luxburg and S. Bengio and H. Wallach and R. Fergus and S. Vishwanathan and R. Garnett},
 pages = {},
 publisher = {Curran Associates, Inc.},
 title = {{Stabilizing Training of Generative Adversarial Networks through Regularization}},
 url = {https://proceedings.neurips.cc/paper_files/paper/2017/file/7bccfde7714a1ebadf06c5f4cea752c1-Paper.pdf},
 volume = {30},
 year = {2017}
}

@inproceedings{Nagarajan17,
 author = {Nagarajan, Vaishnavh and Kolter, J. Zico},
 booktitle = {Advances in Neural Information Processing Systems},
 editor = {I. Guyon and U. Von Luxburg and S. Bengio and H. Wallach and R. Fergus and S. Vishwanathan and R. Garnett},
 pages = {},
 publisher = {Curran Associates, Inc.},
 title = {Gradient descent {GAN} optimization is locally stable},
 volume = {30},
 year = {2017}
}

@inproceedings{Foret21,
title={{Sharpness-Aware Minimization for Efficiently Improving Generalization}},
author={Pierre Foret and Ariel Kleiner and Hossein Mobahi and Behnam Neyshabur},
booktitle={International Conference on Learning Representations},
year={2021},
url={https://openreview.net/forum?id=6Tm1mposlrM}
}

@inproceedings{Arjovsky17b,
title={Towards {P}rincipled {M}ethods for {T}raining {G}enerative {A}dversarial {N}etworks},
author={Martin Arjovsky and Leon Bottou},
booktitle={International Conference on Learning Representations},
year={2017},
url={https://openreview.net/forum?id=Hk4_qw5xe}
}

@InProceedings{Arjovsky17,
  title = 	 {{W}asserstein {G}enerative {A}dversarial {N}etworks},
  author =       {Martin Arjovsky and Soumith Chintala and L{\'e}on Bottou},
  booktitle = 	 {Proceedings of the 34th International Conference on Machine Learning},
  pages = 	 {214--223},
  year = 	 {2017},
  editor = 	 {Precup, Doina and Teh, Yee Whye},
  volume = 	 {70},
  series = 	 {Proceedings of Machine Learning Research},
  month = 	 {06--11 Aug},
  publisher =    {PMLR},
}

@InProceedings{Zheng23,
  title = 	 {{Fast Sampling of Diffusion Models via Operator Learning}},
  author =       {Zheng, Hongkai and Nie, Weili and Vahdat, Arash and Azizzadenesheli, Kamyar and Anandkumar, Anima},
  booktitle = 	 {Proceedings of the 40th International Conference on Machine Learning},
  pages = 	 {42390--42402},
  year = 	 {2023},
  editor = 	 {Krause, Andreas and Brunskill, Emma and Cho, Kyunghyun and Engelhardt, Barbara and Sabato, Sivan and Scarlett, Jonathan},
  volume = 	 {202},
  series = 	 {Proceedings of Machine Learning Research},
  month = 	 {23--29 Jul},
  publisher =    {PMLR},
}

@InProceedings{Sharma23,
  title = 	 {{Do {B}ayesian Neural Networks Need To Be Fully Stochastic?}},
  author =       {Sharma, Mrinank and Farquhar, Sebastian and Nalisnick, Eric and Rainforth, Tom},
  booktitle = 	 {Proceedings of The 26th International Conference on Artificial Intelligence and Statistics},
  pages = 	 {7694--7722},
  year = 	 {2023},
  editor = 	 {Ruiz, Francisco and Dy, Jennifer and van de Meent, Jan-Willem},
  volume = 	 {206},
  series = 	 {Proceedings of Machine Learning Research},
  month = 	 {25--27 Apr},
  publisher =    {PMLR},
  url = 	 {https://proceedings.mlr.press/v206/sharma23a.html},
}

@inproceedings{Akaike73,
  title        = {{Information Theory and an Extension of the Maximum Likelihood Principle}},
  author       = {Akaike, H},
  booktitle    = {2nd International Symposium on Information Theory, 1973},
  pages        = {268--281},
  year         = {1973},
  organization = {Publishing House of the Hungarian Academy of Sciences}
}

@inproceedings{Wang23,
  title     = {{Diffusion-{GAN}: Training {GAN}s with Diffusion}},
  author    = {Zhendong Wang and Huangjie Zheng and Pengcheng He and Weizhu Chen and Mingyuan Zhou},
  booktitle = {International Conference on Learning Representations },
  year      = {2023}
}

@article{LoaizaGanem22,
  title   = {{Diagnosing and Fixing Manifold Overfitting in Deep Generative Models}},
  author  = {Gabriel Loaiza-Ganem and Brendan Leigh Ross and Jesse C Cresswell and Anthony L. Caterini},
  journal = {Transactions on Machine Learning Research},
  issn    = {2835-8856},
  year    = {2022},
  note    = {}
}

@inproceedings{Brown23,
  title     = {{Verifying the Union of Manifolds Hypothesis for Image Data}},
  author    = {Bradley CA Brown and Anthony L. Caterini and Brendan Leigh Ross and Jesse C Cresswell and Gabriel Loaiza-Ganem},
  booktitle = {International Conference on Learning Representations },
  year      = {2023}
}

@article{Gretton12,
  author  = {Arthur Gretton and Karsten M. Borgwardt and Malte J. Rasch and Bernhard Sch{{\"o}}lkopf and Alexander Smola},
  title   = {{A Kernel Two-Sample Test}},
  journal = {Journal of Machine Learning Research},
  year    = {2012},
  volume  = {13},
  number  = {25},
  pages   = {723-773}
}

@inproceedings{TranNeurIPS21,
 author = {Tran, Ba-Hien and Rossi, Simone and Milios, Dimitrios and Michiardi, Pietro and Bonilla, Edwin V and Filippone, Maurizio},
 booktitle = {Advances in Neural Information Processing Systems},
 editor = {M. Ranzato and A. Beygelzimer and Y. Dauphin and P.S. Liang and J. Wortman Vaughan},
 pages = {19730--19742},
 publisher = {Curran Associates, Inc.},
 title = {{Model Selection for {B}ayesian Autoencoders}},
 url = {https://proceedings.neurips.cc/paper_files/paper/2021/file/a41db61e2728ef963614a8c8755b9b9a-Paper.pdf},
 volume = {34},
 year = {2021}
}

@InProceedings{Li15,
  title = 	 {{Generative Moment Matching Networks}},
  author = 	 {Li, Yujia and Swersky, Kevin and Zemel, Rich},
  booktitle = 	 {Proceedings of the 32nd International Conference on Machine Learning},
  pages = 	 {1718--1727},
  year = 	 {2015},
  editor = 	 {Bach, Francis and Blei, David},
  volume = 	 {37},
  series = 	 {Proceedings of Machine Learning Research},
  address = 	 {Lille, France},
  month = 	 {07--09 Jul},
  publisher =    {PMLR},
  url = 	 {https://proceedings.mlr.press/v37/li15.html},
}

@inproceedings{Li17,
 author = {Li, Chun-Liang and Chang, Wei-Cheng and Cheng, Yu and Yang, Yiming and Poczos, Barnabas},
 booktitle = {Advances in Neural Information Processing Systems},
 editor = {I. Guyon and U. Von Luxburg and S. Bengio and H. Wallach and R. Fergus and S. Vishwanathan and R. Garnett},
 pages = {},
 publisher = {Curran Associates, Inc.},
 title = {{MMD {GAN}: Towards Deeper Understanding of Moment Matching Network}},
 url = {https://proceedings.neurips.cc/paper_files/paper/2017/file/dfd7468ac613286cdbb40872c8ef3b06-Paper.pdf},
 volume = {30},
 year = {2017}
}

@inproceedings{Dziugaite15,
author = {Dziugaite, Gintare Karolina and Roy, Daniel M. and Ghahramani, Zoubin},
title = {Training generative neural networks via {M}aximum {M}ean {D}iscrepancy optimization},
year = {2015},
isbn = {9780996643108},
publisher = {AUAI Press},
address = {Arlington, Virginia, USA},
booktitle = {Proceedings of the Thirty-First Conference on Uncertainty in Artificial Intelligence},
pages = {258–267},
numpages = {10},
location = {Amsterdam, Netherlands},
series = {UAI'15}
}

@inproceedings{Nowozin16,
 author = {Nowozin, Sebastian and Cseke, Botond and Tomioka, Ryota},
 booktitle = {Advances in Neural Information Processing Systems},
 editor = {D. Lee and M. Sugiyama and U. Luxburg and I. Guyon and R. Garnett},
 pages = {},
 publisher = {Curran Associates, Inc.},
 title = {{f-{GAN}: Training Generative Neural Samplers using Variational Divergence Minimization}},
 url = {https://proceedings.neurips.cc/paper_files/paper/2016/file/cedebb6e872f539bef8c3f919874e9d7-Paper.pdf},
 volume = {29},
 year = {2016}
}

@inproceedings{Goodfellow14,
 author = {Goodfellow, Ian and Pouget-Abadie, Jean and Mirza, Mehdi and Xu, Bing and Warde-Farley, David and Ozair, Sherjil and Courville, Aaron and Bengio, Yoshua},
 booktitle = {Advances in Neural Information Processing Systems},
 editor = {Z. Ghahramani and M. Welling and C. Cortes and N. Lawrence and K.Q. Weinberger},
 pages = {},
 publisher = {Curran Associates, Inc.},
 title = {{Generative Adversarial Nets}},
 url = {https://proceedings.neurips.cc/paper_files/paper/2014/file/5ca3e9b122f61f8f06494c97b1afccf3-Paper.pdf},
 volume = {27},
 year = {2014}
}

@InProceedings{Fatras20,
  title = 	 {Learning with minibatch {W}asserstein  : asymptotic and gradient properties},
  author =       {Fatras, Kilian and Zine, Younes and Flamary, R\'emi and Gribonval, Remi and Courty, Nicolas},
  booktitle = 	 {Proceedings of the Twenty Third International Conference on Artificial Intelligence and Statistics},
  pages = 	 {2131--2141},
  year = 	 {2020},
  editor = 	 {Chiappa, Silvia and Calandra, Roberto},
  volume = 	 {108},
  series = 	 {Proceedings of Machine Learning Research},
  month = 	 {26--28 Aug},
  publisher =    {PMLR},
  url = 	 {https://proceedings.mlr.press/v108/fatras20a.html},
}

@article{Tiao18,
  author       = {Louis C. Tiao and
                  Edwin V. Bonilla and
                  Fabio Ramos},
  title        = {{Cycle-Consistent Adversarial Learning as Approximate Bayesian Inference}},
  journal      = {CoRR},
  volume       = {abs/1806.01771},
  year         = {2018},
  url          = {http://arxiv.org/abs/1806.01771},
  eprinttype    = {arXiv},
  eprint       = {1806.01771},
  bibsource    = {dblp computer science bibliography, https://dblp.org}
}

@INPROCEEDINGS{Zhu17,
  author={Zhu, Jun-Yan and Park, Taesung and Isola, Phillip and Efros, Alexei A.},
  booktitle={2017 IEEE International Conference on Computer Vision (ICCV)}, 
  title={{Unpaired Image-to-Image Translation Using Cycle-Consistent Adversarial Networks}}, 
  year={2017},
  volume={},
  number={},
  pages={2242-2251},
  doi={10.1109/ICCV.2017.244}
  }

@InProceedings{Mescheder17,
  title = 	 {Adversarial {V}ariational {B}ayes: {U}nifying {V}ariational {A}utoencoders and {G}enerative {A}dversarial {N}etworks},
  author =       {Lars Mescheder and Sebastian Nowozin and Andreas Geiger},
  booktitle = 	 {Proceedings of the 34th International Conference on Machine Learning},
  pages = 	 {2391--2400},
  year = 	 {2017},
  editor = 	 {Precup, Doina and Teh, Yee Whye},
  volume = 	 {70},
  series = 	 {Proceedings of Machine Learning Research},
  month = 	 {06--11 Aug},
  publisher =    {PMLR},
  url = 	 {https://proceedings.mlr.press/v70/mescheder17a.html},
}

@inproceedings{Saatchi17,
 author = {Saatci, Yunus and Wilson, Andrew G},
 booktitle = {Advances in Neural Information Processing Systems},
 editor = {I. Guyon and U. Von Luxburg and S. Bengio and H. Wallach and R. Fergus and S. Vishwanathan and R. Garnett},
 pages = {},
 publisher = {Curran Associates, Inc.},
 title = {{Bayesian GAN}},
 url = {https://proceedings.neurips.cc/paper_files/paper/2017/file/312351bff07989769097660a56395065-Paper.pdf},
 volume = {30},
 year = {2017}
}

@inproceedings{Gretton06,
 author = {Gretton, Arthur and Borgwardt, Karsten and Rasch, Malte and Sch\"{o}lkopf, Bernhard and Smola, Alex},
 booktitle = {Advances in Neural Information Processing Systems},
 editor = {B. Sch\"{o}lkopf and J. Platt and T. Hoffman},
 pages = {},
 publisher = {MIT Press},
 title = {{A Kernel Method for the Two-Sample-Problem}},
 url = {https://proceedings.neurips.cc/paper_files/paper/2006/file/e9fb2eda3d9c55a0d89c98d6c54b5b3e-Paper.pdf},
 volume = {19},
 year = {2006}
}

@ARTICLE{MacKay95,
       author = {{MacKay}, David J.~C.},
        title = "{Bayesian neural networks and density networks}",
      journal = {Nuclear Instruments and Methods in Physics Research A},
         year = 1995,
        month = feb,
       volume = {354},
       number = {1},
        pages = {73-80},
          doi = {10.1016/0168-9002(94)00931-7},
       adsurl = {https://ui.adsabs.harvard.edu/abs/1995NIMPA.354...73M}
}

@ARTICLE{Bishop98,
  author={Bishop, Christopher M. and Svensén, Markus and Williams, Christopher K. I.},
  journal={Neural Computation}, 
  title={{GTM: The Generative Topographic Mapping}}, 
  year={1998},
  volume={10},
  number={1},
  pages={215-234},
  doi={10.1162/089976698300017953}}

@inproceedings{Nguyen07,
 author = {Nguyen, XuanLong and Wainwright, Martin J and Jordan, Michael},
 booktitle = {Advances in Neural Information Processing Systems},
 editor = {J. Platt and D. Koller and Y. Singer and S. Roweis},
 pages = {},
 publisher = {Curran Associates, Inc.},
 title = {Estimating divergence functionals and the likelihood ratio by penalized convex risk minimization},
 url = {https://proceedings.neurips.cc/paper_files/paper/2007/file/72da7fd6d1302c0a159f6436d01e9eb0-Paper.pdf},
 volume = {20},
 year = {2007}
}

@book{Villani16,
  title={Optimal Transport: Old and New},
  author={Villani, C.},
  isbn={9783662501801},
  series={Grundlehren der mathematischen Wissenschaften},
  url={https://books.google.com.sa/books?id=5p8SDAEACAAJ},
  year={2016},
  publisher={Springer Berlin Heidelberg}
}

@article{TranJMLR22,
  author  = {Ba-Hien Tran and Simone Rossi and Dimitrios Milios and Maurizio Filippone},
  title   = {All {Y}ou {N}eed is a {G}ood {F}unctional {P}rior for {B}ayesian {D}eep {L}earning},
  journal = {Journal of Machine Learning Research},
  year    = {2022},
  volume  = {23},
  number  = {74},
  pages   = {1--56},
  url     = {http://jmlr.org/papers/v23/20-1340.html}
}

@inproceedings{FlamShepherd17,
  author    = {Flam-Shepherd, Daniel and Requeima, James and Duvenaud, David},
  booktitle = {NeurIPS workshop on Bayesian Deep Learning},
  title     = {{Mapping Gaussian Process Priors to Bayesian Neural Networks}},
  year      = {2017}
}

@InProceedings{Balaji19,
  title = 	 {Entropic {GAN}s meet {VAE}s: A {S}tatistical {A}pproach to {C}ompute {S}ample {L}ikelihoods in {GAN}s},
  author =       {Balaji, Yogesh and Hassani, Hamed and Chellappa, Rama and Feizi, Soheil},
  booktitle = 	 {Proceedings of the 36th International Conference on Machine Learning},
  pages = 	 {414--423},
  year = 	 {2019},
  editor = 	 {Chaudhuri, Kamalika and Salakhutdinov, Ruslan},
  volume = 	 {97},
  series = 	 {Proceedings of Machine Learning Research},
  month = 	 {09--15 Jun},
  publisher =    {PMLR},
  url = 	 {https://proceedings.mlr.press/v97/balaji19a.html}
}

@inproceedings{
Liu19,
title={{DARTS}: {D}ifferentiable {A}rchitecture {S}earch},
author={Hanxiao Liu and Karen Simonyan and Yiming Yang},
booktitle={International Conference on Learning Representations},
year={2019},
url={https://openreview.net/forum?id=S1eYHoC5FX},
}

@article{Solomonoff64,
title = {A formal theory of inductive inference. {Part I}},
journal = {Information and Control},
volume = {7},
number = {1},
pages = {1-22},
year = {1964},
issn = {0019-9958},
doi = {https://doi.org/10.1016/S0019-9958(64)90223-2},
url = {https://www.sciencedirect.com/science/article/pii/S0019995864902232},
author = {R.J. Solomonoff},
}

@article{muller1997integral, title={{Integral Probability Metrics and Their Generating Classes of Functions}}, volume={29}, DOI={10.2307/1428011}, number={2}, journal={Advances in Applied Probability}, author={M\"uller, Alfred}, year={1997}, pages={429–443}}

@article{Schmidhuber90,
  author       = {J{\"{u}}rgen Schmidhuber},
  title        = {Making the world differentiable: on using self supervised fully recurrent
                  neural networks for dynamic reinforcement learning and planning in
                  non-stationary environments},
  journal      = {Forschungsberichte, {TU} Munich},
  volume       = {{FKI} 126 90},
  pages        = {1--26},
  year         = {1990},
  url          = {https://d-nb.info/920771238},
  bibsource    = {dblp computer science bibliography, https://dblp.org}
}

@inproceedings{Schmidhuber91, author = {Schmidhuber, J\"{u}rgen}, title = {{A Possibility for Implementing Curiosity and Boredom in
Model-Building Neural Controllers}}, year = {1991}, isbn = {0262631385}, publisher = {MIT Press}, address = {Cambridge, MA, USA}, booktitle = {Proceedings of the First International Conference on Simulation of Adaptive Behavior on From Animals to Animats}, pages = {222–227}, numpages = {6}, location = {Paris, France} }

@inproceedings{Schmidhuber95, author = {Schmidhuber, J\"{u}rgen}, title = {{Discovering Solutions with Low Kolmogorov Complexity and High
Generalization Capability}}, year = {1995}, isbn = {1558603778}, publisher = {Morgan Kaufmann Publishers Inc.}, address = {San Francisco, CA, USA}, booktitle = {Proceedings of the Twelfth International Conference on International Conference on Machine Learning}, pages = {488–496}, numpages = {9}, location = {Tahoe City, California, USA}, series = {ICML'95} }

@article{Hochreiter97,
    author = {Hochreiter, Sepp and Schmidhuber, Jürgen},
    title = {Flat {M}inima},
    journal = {Neural Computation},
    volume = {9},
    number = {1},
    pages = {1-42},
    year = {1997},
    month = {01},
    issn = {0899-7667},
    doi = {10.1162/neco.1997.9.1.1},
    url = {https://doi.org/10.1162/neco.1997.9.1.1},
    eprint = {https://direct.mit.edu/neco/article-pdf/9/1/1/813385/neco.1997.9.1.1.pdf},
}

@inproceedings{Wang23b, author = {Wang, Yite and Wu, Jing and Hovakimyan, Naira and Sun, Ruoyu}, title = {{Balanced Training for Sparse {GAN}s}}, year = {2023}, publisher = {Curran Associates Inc.}, address = {Red Hook, NY, USA}, booktitle = {Proceedings of the 37th International Conference on Neural Information Processing Systems}, articleno = {600}, numpages = {24}, location = {New Orleans, LA, USA}, series = {NIPS '23} }

@inproceedings{Gulrajani17,
 author = {Gulrajani, Ishaan and Ahmed, Faruk and Arjovsky, Martin and Dumoulin, Vincent and Courville, Aaron C},
 booktitle = {Advances in Neural Information Processing Systems},
 editor = {I. Guyon and U. Von Luxburg and S. Bengio and H. Wallach and R. Fergus and S. Vishwanathan and R. Garnett},
 pages = {},
 publisher = {Curran Associates, Inc.},
 title = {Improved {T}raining of {W}asserstein {GAN}s},
 url = {https://proceedings.neurips.cc/paper_files/paper/2017/file/892c3b1c6dccd52936e27cbd0ff683d6-Paper.pdf},
 volume = {30},
 year = {2017}
}

@inproceedings{Gal16,
    author = {Gal, Yarin and Ghahramani, Zoubin},
    booktitle = {Proceedings of the 33rd International Conference on International Conference on Machine Learning - Volume 48},
    location = {New York, NY, USA},
    pages = {1050--1059},
    publisher = {JMLR.org},
    series = {ICML'16},
    title = {{Dropout As a Bayesian Approximation: Representing Model Uncertainty in Deep Learning}},
    url = {http://dl.acm.org/citation.cfm?id=3045390.3045502},
    year = {2016}
}

@article{Gneiting07,
    author = {Gneiting, Tilmann and Raftery, Adrian E.},
    journal = {Journal of the American Statistical Association},
    pages = {359--378},
    title = {{Strictly Proper Scoring Rules, Prediction, and Estimation}},
    volume = {102},
    year = {2007}
}

@book{Neal96,
    author = {Neal, Radford M.},
    day = {09},
    edition = {1},
    howpublished = {Paperback},
    isbn = {0387947248},
    month = aug,
    publisher = {Springer},
    title = {{Bayesian Learning for Neural Networks (Lecture Notes in Statistics)}},
    url = {http://www.worldcat.org/isbn/0387947248},
    year = {1996}
}

@incollection{Mackay94,
    author = {Mac{k}ay, D. J. C.},
    booktitle = {Models of Neural Networks {III}},
    chapter = {6},
    editor = {Domany, E. and van Hemmen, J. L. and Schulten, K.},
    pages = {211--254},
    publisher = {Springer},
    title = {{B}ayesian {M}ethods for {B}ackpropagation {N}etworks},
    year = {1994}
}

@manual{R,
    address = {Vienna, Austria},
    author = {{R Development Core Team}},
    comment = {{ISBN} 3-900051-07-0},
    organization = {R Foundation for Statistical Computing},
    title = {{R: A Language and Environment for Statistical Computing}},
    url = {http://www.R-project.org},
    year = {2006}
}

@article{DBLP:journals/corr/abs-1812-04948,
  publtype={informal},
  author={Tero Karras and Samuli Laine and Timo Aila},
  title={{A Style-Based Generator Architecture for Generative Adversarial Networks}},
  year={2018},
  cdate={1514764800000},
  journal={CoRR},
  volume={abs/1812.04948},
  url={http://arxiv.org/abs/1812.04948}
}

@INPROCEEDINGS {9156570,
author = { Karras, Tero and Laine, Samuli and Aittala, Miika and Hellsten, Janne and Lehtinen, Jaakko and Aila, Timo },
booktitle = { 2020 IEEE/CVF Conference on Computer Vision and Pattern Recognition (CVPR) },
title = {{ Analyzing and Improving the Image Quality of StyleGAN }},
year = {2020},
volume = {},
ISSN = {},
pages = {8107-8116},
doi = {10.1109/CVPR42600.2020.00813},
url = {https://doi.ieeecomputersociety.org/10.1109/CVPR42600.2020.00813},
publisher = {IEEE Computer Society},
address = {Los Alamitos, CA, USA},
month =Jun}

@inproceedings{NEURIPS2021_076ccd93,
 author = {Karras, Tero and Aittala, Miika and Laine, Samuli and H\"{a}rk\"{o}nen, Erik and Hellsten, Janne and Lehtinen, Jaakko and Aila, Timo},
 booktitle = {Advances in Neural Information Processing Systems},
 editor = {M. Ranzato and A. Beygelzimer and Y. Dauphin and P.S. Liang and J. Wortman Vaughan},
 pages = {852--863},
 publisher = {Curran Associates, Inc.},
 title = {{Alias-Free Generative Adversarial Networks}},
 url = {https://proceedings.neurips.cc/paper_files/paper/2021/file/076ccd93ad68be51f23707988e934906-Paper.pdf},
 volume = {34},
 year = {2021}
}

@inproceedings{NEURIPS2020_8d30aa96,
 author = {Karras, Tero and Aittala, Miika and Hellsten, Janne and Laine, Samuli and Lehtinen, Jaakko and Aila, Timo},
 booktitle = {Advances in Neural Information Processing Systems},
 editor = {H. Larochelle and M. Ranzato and R. Hadsell and M.F. Balcan and H. Lin},
 pages = {12104--12114},
 publisher = {Curran Associates, Inc.},
 title = {{Training Generative Adversarial Networks with Limited Data}},
 url = {https://proceedings.neurips.cc/paper_files/paper/2020/file/8d30aa96e72440759f74bd2306c1fa3d-Paper.pdf},
 volume = {33},
 year = {2020}
}

@InProceedings{pmlr-v115-nie20a,
  title = 	 {Towards a {B}etter {U}nderstanding and {R}egularization of {GAN} {T}raining {D}ynamics},
  author =       {Nie, Weili and Patel, Ankit B.},
  booktitle = 	 {Proceedings of The 35th Uncertainty in Artificial Intelligence Conference},
  pages = 	 {281--291},
  year = 	 {2020},
  editor = 	 {Adams, Ryan P. and Gogate, Vibhav},
  volume = 	 {115},
  series = 	 {Proceedings of Machine Learning Research},
  month = 	 {22--25 Jul},
  publisher =    {PMLR},
  url = 	 {https://proceedings.mlr.press/v115/nie20a.html}
}

@inproceedings{NEURIPS2024_4e2acb1e,
 author = {Huang, Yiwen and Gokaslan, Aaron and Kuleshov, Volodymyr and Tompkin, James},
 booktitle = {Advances in Neural Information Processing Systems},
 editor = {A. Globerson and L. Mackey and D. Belgrave and A. Fan and U. Paquet and J. Tomczak and C. Zhang},
 pages = {44177--44215},
 publisher = {Curran Associates, Inc.},
 title = {The {GAN} is dead; long live the {GAN}! {A} {M}odern {GAN} {B}aeline},
 url = {https://proceedings.neurips.cc/paper_files/paper/2024/file/4e2acb1e1c8e297d394ae29ed9535172-Paper-Conference.pdf},
 volume = {37},
 year = {2024}
}

@InProceedings{pmlr-v119-farnia20a,
  title = 	 {Do {GAN}s always have {N}ash equilibria?},
  author =       {Farnia, Farzan and Ozdaglar, Asuman},
  booktitle = 	 {Proceedings of the 37th International Conference on Machine Learning},
  pages = 	 {3029--3039},
  year = 	 {2020},
  editor = 	 {III, Hal Daumé and Singh, Aarti},
  volume = 	 {119},
  series = 	 {Proceedings of Machine Learning Research},
  month = 	 {13--18 Jul},
  publisher =    {PMLR},
  url = 	 {https://proceedings.mlr.press/v119/farnia20a.html}
}

@inproceedings{NEURIPS2020_a851bd0d,
 author = {Sinha, Samarth and Zhao, Zhengli and Goyal, Anirudh and Raffel, Colin A and Odena, Augustus},
 booktitle = {Advances in Neural Information Processing Systems},
 editor = {H. Larochelle and M. Ranzato and R. Hadsell and M.F. Balcan and H. Lin},
 pages = {14638--14649},
 publisher = {Curran Associates, Inc.},
 title = {{Top-k Training of {GAN}s: Improving {GAN} Performance by Throwing Away Bad Samples}},
 url = {https://proceedings.neurips.cc/paper_files/paper/2020/file/a851bd0d418b13310dd1e5e3ac7318ab-Paper.pdf},
 volume = {33},
 year = {2020}
}

@inproceedings{
jolicoeur-martineau2018,
title={ The relativistic discriminator: a key element missing from standard {GAN}},
author={Alexia Jolicoeur-Martineau},
booktitle={International Conference on Learning Representations},
year={2019},
url={https://openreview.net/forum?id=S1erHoR5t7},
}
\bibliographystyle{icml2026_nourl}

\newpage
\appendix
\onecolumn

\section{Experiments using a larger batch size}

In this section, we report results using a larger batch size. 
The results show poorer performance, but again the best configurations are associated with some form or regularization possibly in combination with \sam. 
Despite the effectiveness of \sam, even for a larger batch-size compared with the experiments in the main paper, flatter minima are generally not associated with better performance. 

\begin{table}[h]
  \caption{Experiments using a larger batch-size $|B|=512$. 
  }
  \label{tab:res:batch_size:mnist_cifar}
  \begin{center}
    \begin{small}
      \begin{sc}
\resizebox{\textwidth}{!}{\begin{tabular}{lll|llll|llll}
    \toprule
    & & & \multicolumn{4}{c|}{\mnist} &  \multicolumn{4}{c}{\cifar} \\
    \midrule    
    \multicolumn{11}{c}{\wgan \quad $|B|=512$} \\ 
    \midrule
  $\sigma^2_{\mathrm{lik}}$  &  $\lambda_{\mathrm{gr}}$  & $\rho_{\mathrm{SAM}}$  &   
    \iscscore $\uparrow$   &  \fidscore $\downarrow$  &  \kidscore {\scriptsize $\, \times 10^{-3} \,$} $\downarrow$  &
    $\ell_1$ & 
    \iscscore $\uparrow$   &  \fidscore $\downarrow$  &  \kidscore {\scriptsize $\, \times 10^{-3} \,$} $\downarrow$  &
    $\ell_1$ \\ 
    \midrule
    0.0	 &0.0	 &0.0	 & 2.18 (0.02) 	 & 17.7 (0.9) 	 & 14.6 (1.0) 	 & 23.6 (10.2) 	     & 3.66 (0.16) 	 & 107.1 (3.9) 	 & 97.5 (4.1) 	 & 10.2 (6.8) 	 \\   
0.01	 &0.0	 &0.0	 & 2.11 (0.12) 	 & 21.2 (5.3) 	 & 18.4 (6.1) 	 & 24.2 (12.4) 	     & 3.55 (0.08) 	 & 107.6 (5.8) 	 & 97.5 (6.3) 	 & 8.1 (3.2) 	 \\   
0.001	 &0.0	 &0.0	 & 2.17 (0.02) 	 & 17.9 (1.0) 	 & 14.8 (1.0) 	 & 22.1 (12.2) 	     & 3.65 (0.13) 	 & 107.7 (3.6) 	 & 97.3 (3.6) 	 & 9.9 (7.4) 	 \\   
0.0	 &0.01	 &0.0	 & 2.17 (0.02) 	 & 18.2 (0.9) 	 & 14.9 (0.9) 	 & 25.3 (10.9) 	     & 3.68 (0.07) 	 & 104.9 (4.6) 	 & 95.5 (5.6) 	 & 11.3 (5.6) 	 \\   
0.01	 &0.01	 &0.0	 & 2.20 (0.02) 	 & 18.9 (0.3) 	 & 15.5 (0.2) 	 & 24.0 (9.5) 	     & 3.65 (0.11) 	 & 105.9 (5.8) 	 & 95.5 (5.8) 	 & 11.8 (9.9) 	 \\   
0.001	 &0.01	 &0.0	 & 2.18 (0.03) 	 & 18.0 (0.5) 	 & 14.9 (0.5) 	 & 22.5 (7.8) 	     & 3.68 (0.16) 	 & 107.8 (4.7) 	 & 97.2 (4.8) 	 & 9.7 (7.8) 	 \\   
0.0	 &0.001	 &0.0	 & 2.18 (0.02) 	 & 18.4 (0.5) 	 & 15.2 (0.6) 	 & 30.7 (8.3) 	     & 3.55 (0.13) 	 & 108.2 (4.4) 	 & 98.3 (4.8) 	 & 16.2 (3.4) 	 \\   
0.01	 &0.001	 &0.0	 & 2.15 (0.08) 	 & 18.5 (1.3) 	 & 15.4 (1.6) 	 & 22.4 (9.2) 	     & 3.65 (0.18) 	 & 104.0 (3.7) 	 & 93.1 (4.4) 	 & 16.6 (6.2) 	 \\   
0.001	 &0.001	 &0.0	 & 2.19 (0.01) 	 & 18.1 (1.0) 	 & 14.9 (1.0) 	 & 22.9 (9.4) 	     & 3.67 (0.09) 	 & 108.1 (2.6) 	 & 97.4 (2.6) 	 & 13.2 (4.7) 	 \\   
0.0	 &0.0	 &0.01	 & 2.17 (0.02) 	 & 18.0 (0.5) 	 & 14.8 (0.5) 	 & 24.7 (9.5) 	     & 3.65 (0.10) 	 & 110.6 (6.8) 	 & 100.6 (7.6) 	 & 18.6 (8.4) 	 \\   
0.01	 &0.0	 &0.01	 & 2.18 (0.02) 	 & 18.3 (0.7) 	 & 15.0 (0.6) 	 & 18.6 (4.9) 	     & 3.62 (0.06) 	 & 108.7 (4.2) 	 & 98.6 (3.6) 	 & 9.6 (2.8) 	 \\   
0.001	 &0.0	 &0.01	 & 2.17 (0.02) 	 & 17.9 (0.7) 	 & 14.8 (0.5) 	 & 24.1 (3.2) 	     & 3.71 (0.13) 	 & 107.3 (2.8) 	 & 97.1 (4.4) 	 & 13.7 (2.7) 	 \\   
0.0	 &0.01	 &0.01	 & 2.19 (0.02) 	 & 18.7 (0.7) 	 & 15.4 (0.8) 	 & 22.5 (7.7) 	     & 3.63 (0.07) 	 & 107.8 (3.0) 	 & 96.5 (3.1) 	 & 20.3 (11.4) 	 \\   
0.01	 &0.01	 &0.01	 & 2.18 (0.03) 	 & 18.0 (0.4) 	 & 14.8 (0.4) 	 & 23.0 (9.6) 	     & 3.69 (0.17) 	 & 106.8 (6.9) 	 & 95.7 (7.6) 	 & 11.2 (10.2) 	 \\   
0.001	 &0.01	 &0.01	 & 2.18 (0.02) 	 & 18.2 (0.8) 	 & 15.0 (0.8) 	 & 25.5 (12.8) 	     & 3.65 (0.08) 	 & 106.5 (6.0) 	 & 96.3 (6.6) 	 & 12.5 (5.8) 	 \\   
0.0	 &0.001	 &0.01	 & 2.16 (0.03) 	 & 17.9 (0.8) 	 & 14.8 (1.0) 	 & 21.0 (9.7) 	     & 3.72 (0.08) 	 & 110.4 (6.6) 	 & 100.1 (6.7) 	 & 15.4 (12.0) 	 \\   
0.01	 &0.001	 &0.01	 & 2.17 (0.02) 	 & 19.0 (2.2) 	 & 15.7 (2.3) 	 & 16.9 (8.6) 	     & 3.52 (0.13) 	 & 110.1 (4.3) 	 & 100.3 (4.6) 	 & 12.3 (6.0) 	 \\   
0.001	 &0.001	 &0.01	 & 2.17 (0.02) 	 & 18.3 (0.7) 	 & 15.1 (0.7) 	 & 23.3 (6.1) 	     & 3.62 (0.11) 	 & 111.9 (8.8) 	 & 101.6 (9.3) 	 & 18.8 (12.4) 	 \\   
0.0	 &0.0	 &0.1	 & 2.18 (0.01) 	 & 17.6 (0.5) 	 & 14.6 (0.5) 	 & 13.6 (2.4) 	     & 3.64 (0.10) 	 & 111.2 (4.9) 	 & 101.3 (4.9) 	 & 16.8 (5.1) 	 \\   
0.01	 &0.0	 &0.1	 & 2.17 (0.02) 	 & 18.1 (0.8) 	 & 15.0 (0.8) 	 & 23.5 (8.5) 	     & 3.69 (0.14) 	 & 109.1 (8.2) 	 & 99.7 (8.9) 	 & 15.8 (6.3) 	 \\   
0.001	 &0.0	 &0.1	 & 2.19 (0.03) 	 & 17.8 (0.4) 	 & 14.6 (0.4) 	 & 16.3 (11.5) 	     & 3.64 (0.10) 	 & 113.4 (6.8) 	 & 103.9 (8.3) 	 & 14.6 (7.8) 	 \\   
0.0	 &0.01	 &0.1	 & 2.17 (0.01) 	 & 17.3 (0.6) 	 & 14.4 (0.6) 	 & 21.7 (9.9) 	     & 3.69 (0.09) 	 & 111.9 (5.8) 	 & 102.6 (6.2) 	 & 11.0 (5.6) 	 \\   
0.01	 &0.01	 &0.1	 & 2.18 (0.01) 	 & 17.7 (1.0) 	 & 14.6 (0.8) 	 & 21.2 (12.1) 	     & 3.64 (0.11) 	 & 111.2 (5.0) 	 & 100.8 (5.6) 	 & 19.2 (8.0) 	 \\   
0.001	 &0.01	 &0.1	 & 2.19 (0.02) 	 & 17.5 (0.6) 	 & 14.2 (0.6) 	 & 23.2 (4.0) 	     & 3.65 (0.12) 	 & 112.2 (5.9) 	 & 103.2 (6.4) 	 & 9.6 (5.3) 	 \\   
0.0	 &0.001	 &0.1	 & 2.18 (0.02) 	 & 17.4 (0.7) 	 & 14.3 (0.8) 	 & 15.3 (4.9) 	     & 3.64 (0.23) 	 & 110.9 (3.7) 	 & 101.1 (4.5) 	 & 17.9 (6.6) 	 \\   
0.01	 &0.001	 &0.1	 & 2.18 (0.02) 	 & 17.5 (0.9) 	 & 14.3 (0.8) 	 & 19.4 (8.7) 	     & 3.67 (0.13) 	 & 110.9 (3.9) 	 & 99.9 (4.2) 	 & 13.8 (6.5) 	 \\   
0.001	 &0.001	 &0.1	 & 2.17 (0.02) 	 & 17.6 (0.5) 	 & 14.5 (0.5) 	 & 19.9 (8.2) 	     & 3.68 (0.14) 	 & 110.2 (2.5) 	 & 100.2 (3.5) 	 & 21.6 (7.4) 	 \\   
    \bottomrule
  \end{tabular}}
  \end{sc}
  \end{small}
  \end{center}
\end{table}

\section{Experiments using \mcd with multiple Monte Carlo samples}

In the experiments in the main paper, at each training iteration, we considered \mcd with one Monte Carlo sample. 
A natural question is what happens when increasing the number of Monte Carlo samples, and as expected performance is generally improved.

\begin{table}[!h]
  \caption{Comparison of \mcd with one or five Monte Carlo samples during training. 
  }
  \label{tab:res:mcd_multiple:mnist}
  \begin{center}
    \begin{small}
      \begin{sc}
\resizebox{\textwidth}{!}{\begin{tabular}{lll|llll|llll}
    \toprule
    & & & \multicolumn{4}{c|}{$N_{\mcd} = 5$} &  \multicolumn{4}{c}{$N_{\mcd} = 1$} \\
    \midrule    
    \multicolumn{11}{c}{\wgan \quad $|B|=128$ \quad \mnist} \\ 
    \midrule
  $\sigma^2_{\mathrm{lik}}$  &  $\lambda_{\mathrm{gr}}$  & $\rho_{\mathrm{SAM}}$  &   
    \iscscore $\uparrow$   &  \fidscore $\downarrow$  &  \kidscore {\scriptsize $\, \times 10^{-3} \,$} $\downarrow$  &
    $\ell_1$ & 
    \iscscore $\uparrow$   &  \fidscore $\downarrow$  &  \kidscore {\scriptsize $\, \times 10^{-3} \,$} $\downarrow$  &
    $\ell_1$ \\ 
    \midrule
0.0	 &0.0	 &0.0	 & 2.23 (0.00) 	 & 7.9 (0.1) 	 & 5.4 (0.1) 	 & 43.9 (0.0) 	   & 2.25 (0.02) 	 & 7.7 (0.5) 	 & 5.4 (0.5) 	 & 20.2 (15.1) 	 \\   
0.01	 &0.0	 &0.0	 & 2.26 (0.01) 	 & 7.6 (0.0) 	 & 5.1 (0.0) 	 & 44.7 (0.0) 	   & 2.26 (0.02) 	 & 7.8 (0.3) 	 & 5.4 (0.3) 	 & 30.5 (11.8) 	 \\   
0.001	 &0.0	 &0.0	 & 2.24 (0.01) 	 & 7.1 (0.0) 	 & 4.8 (0.1) 	 & 39.2 (0.0) 	   & 2.24 (0.01) 	 & 8.0 (0.3) 	 & 5.6 (0.4) 	 & 34.1 (8.3) 	 \\   
0.0	 &0.01	 &0.0	 & 2.23 (0.02) 	 & 7.8 (0.1) 	 & 5.5 (0.2) 	 & 31.1 (0.0) 	   & 2.25 (0.01) 	 & 8.1 (0.8) 	 & 5.7 (0.8) 	 & 19.7 (8.7) 	 \\   
0.01	 &0.01	 &0.0	 & 2.24 (0.01) 	 & 8.0 (0.2) 	 & 5.7 (0.3) 	 & 47.3 (0.0) 	   & 2.26 (0.02) 	 & 7.8 (0.3) 	 & 5.4 (0.3) 	 & 28.3 (11.4) 	 \\   
0.001	 &0.01	 &0.0	 & 2.23 (0.00) 	 & 7.1 (0.1) 	 & 4.7 (0.0) 	 & 43.0 (0.0) 	   & 2.25 (0.02) 	 & 7.8 (0.2) 	 & 5.5 (0.3) 	 & 31.6 (10.0) 	 \\   
0.0	 &0.001	 &0.0	 & 2.26 (0.02) 	 & 7.3 (0.1) 	 & 5.0 (0.1) 	 & 45.0 (0.0) 	   & 2.24 (0.01) 	 & 7.7 (0.4) 	 & 5.5 (0.4) 	 & 31.7 (10.9) 	 \\   
0.01	 &0.001	 &0.0	 & 2.26 (0.01) 	 & 7.4 (0.1) 	 & 5.1 (0.1) 	 & 34.0 (0.0) 	   & 2.24 (0.01) 	 & 8.1 (0.4) 	 & 5.8 (0.4) 	 & 33.3 (7.2) 	 \\   
0.001	 &0.001	 &0.0	 & 2.22 (0.01) 	 & 7.4 (0.1) 	 & 5.1 (0.2) 	 & 34.4 (0.0) 	   & 2.25 (0.01) 	 & 8.0 (0.7) 	 & 5.7 (0.7) 	 & 32.5 (8.4) 	 \\   
    \bottomrule
  \end{tabular}}
  \end{sc}
  \end{small}
  \end{center}
\end{table}

\section{Full set of results on \wgans}

In this section we report all results obtained by combining regularization strategies, \sam, and \mcd on \wgans. 
Even for this model, we found that regularization and \sam optimization generally lead to better performance. 

\begin{table}[!h]
  \caption{\dcgan architecture with Wasserstein divergence objective \citep{Wu18} on \mnist and \cifar.
    Standard deviations, calculated over five seeds times three repetitions of sampling $10\,000$ images, are reported in parenthesis.
  }
  \label{tab:res:all_mnist_cifar}
  \begin{center}
    \begin{small}
      \begin{sc}
\resizebox{\textwidth}{!}{\begin{tabular}{lll|llll|llll}
    \toprule
    & & & \multicolumn{4}{c|}{\mnist} &  \multicolumn{4}{c}{\cifar} \\
    \midrule    
    \multicolumn{11}{c}{\wgan \quad $|B|=128$} \\ 
    \midrule
  $\sigma^2_{\mathrm{lik}}$  &  $\lambda_{\mathrm{gr}}$  & $\rho_{\mathrm{SAM}}$  &   
    \iscscore $\uparrow$   &  \fidscore $\downarrow$  &  \kidscore {\scriptsize $\, \times 10^{-3} \,$} $\downarrow$  &
    $\ell_1$ & 
    \iscscore $\uparrow$   &  \fidscore $\downarrow$  &  \kidscore {\scriptsize $\, \times 10^{-3} \,$} $\downarrow$  &
    $\ell_1$ \\ 
    \midrule
0.0	 &0.0	 &0.0	 & 2.27 (0.02) 	 & 8.4 (0.4) 	 & 6.0 (0.4) 	 & 61.6 (8.3) 	    & 4.64 (0.13) 	 & 40.1 (1.5) 	 & 30.3 (1.6) 	 & 11.6 (5.9) 	 \\
0.01	 &0.0	 &0.0	 & 2.27 (0.02) 	 & 8.0 (0.4) 	 & 5.5 (0.4) 	 & 55.5 (15.1) 	    & 4.74 (0.08) 	 & 34.1 (2.2) 	 & 24.1 (2.1) 	 & 11.2 (3.6) 	 \\
0.001	 &0.0	 &0.0	 & 2.27 (0.02) 	 & 8.3 (0.3) 	 & 5.9 (0.3) 	 & 68.8 (13.5) 	    & 4.72 (0.11) 	 & 37.2 (2.8) 	 & 27.4 (2.6) 	 & 6.6 (3.8) 	 \\
0.0	 &0.01	 &0.0	 & 2.26 (0.01) 	 & 8.2 (0.4) 	 & 5.8 (0.4) 	 & 59.4 (15.4) 	    & 4.72 (0.08) 	 & 37.8 (1.3) 	 & 27.9 (1.6) 	 & 4.6 (3.3) 	 \\
0.01	 &0.01	 &0.0	 & 2.26 (0.01) 	 & 8.3 (0.3) 	 & 5.9 (0.3) 	 & 67.0 (11.1) 	    & 4.79 (0.11) 	 & 34.2 (1.7) 	 & 24.2 (1.9) 	 & 12.3 (4.2) 	 \\
0.001	 &0.01	 &0.0	 & 2.25 (0.01) 	 & 8.3 (0.3) 	 & 6.0 (0.3) 	 & 61.5 (10.7) 	    & 4.71 (0.07) 	 & 37.4 (1.9) 	 & 27.6 (2.1) 	 & 9.8 (4.0) 	 \\
0.0	 &0.001	 &0.0	 & 2.27 (0.02) 	 & 8.0 (0.2) 	 & 5.5 (0.2) 	 & 61.7 (14.3) 	    & 4.68 (0.14) 	 & 40.3 (3.3) 	 & 30.7 (3.4) 	 & 10.2 (5.1) 	 \\
0.01	 &0.001	 &0.0	 & 2.27 (0.02) 	 & 8.4 (0.5) 	 & 6.0 (0.4) 	 & 58.4 (11.8) 	    & 4.76 (0.11) 	 & 33.8 (3.8) 	 & 23.7 (4.1) 	 & 9.1 (5.4) 	 \\
0.001	 &0.001	 &0.0	 & 2.26 (0.01) 	 & 8.1 (0.4) 	 & 5.7 (0.4) 	 & 8.6 (22.3) 	    & 4.76 (0.11) 	 & 40.0 (1.9) 	 & 30.3 (2.1) 	 & 13.4 (4.9) 	 \\
0.0	 &0.0	 &0.01	 & 2.25 (0.01) 	 & 8.4 (0.2) 	 & 6.0 (0.2) 	 & 62.4 (5.4) 	    & 4.70 (0.09) 	 & 39.2 (2.2) 	 & 29.4 (2.4) 	 & 8.7 (6.5) 	 \\
0.01	 &0.0	 &0.01	 & 2.27 (0.02) 	 & 8.0 (0.2) 	 & 5.5 (0.3) 	 & 58.0 (3.2) 	    & 4.74 (0.08) 	 & 36.7 (2.2) 	 & 26.5 (2.2) 	 & 10.4 (2.0) 	 \\
0.001	 &0.0	 &0.01	 & 2.26 (0.02) 	 & 8.0 (0.2) 	 & 5.6 (0.3) 	 & 57.5 (16.1) 	    & 4.70 (0.11) 	 & 41.7 (3.8) 	 & 32.3 (4.1) 	 & 15.6 (2.6) 	 \\
0.0	 &0.01	 &0.01	 & 2.27 (0.02) 	 & 8.1 (0.3) 	 & 5.7 (0.4) 	 & 69.9 (16.8) 	    & 4.79 (0.10) 	 & 39.6 (2.8) 	 & 30.1 (2.9) 	 & 9.5 (3.1) 	 \\
0.01	 &0.01	 &0.01	 & 2.27 (0.02) 	 & 8.3 (0.5) 	 & 5.8 (0.5) 	 & 72.8 (22.6) 	    & 4.79 (0.10) 	 & 33.8 (2.4) 	 & 23.5 (2.3) 	 & 9.7 (3.6) 	 \\
0.001	 &0.01	 &0.01	 & 2.26 (0.01) 	 & 8.5 (0.5) 	 & 6.1 (0.5) 	 & 65.2 (11.6) 	    & 4.74 (0.10) 	 & 39.7 (2.8) 	 & 29.9 (2.6) 	 & 6.7 (4.1) 	 \\
0.0	 &0.001	 &0.01	 & 2.27 (0.02) 	 & 8.1 (0.4) 	 & 5.7 (0.4) 	 & 64.2 (12.1) 	    & 4.76 (0.08) 	 & 38.7 (2.9) 	 & 28.9 (3.1) 	 & 9.9 (4.2) 	 \\
0.01	 &0.001	 &0.01	 & 2.26 (0.01) 	 & 8.2 (0.3) 	 & 5.8 (0.3) 	 & 63.6 (6.3) 	    & 4.78 (0.07) 	 & 33.2 (1.1) 	 & 23.1 (1.3) 	 & 9.6 (2.0) 	 \\
0.001	 &0.001	 &0.01	 & 2.26 (0.02) 	 & 8.3 (0.3) 	 & 5.9 (0.3) 	 & 58.4 (8.5) 	    & 4.65 (0.16) 	 & 41.9 (5.1) 	 & 32.1 (5.2) 	 & 7.9 (2.7) 	 \\
0.0	 &0.0	 &0.1	 & 2.25 (0.02) 	 & 8.6 (0.4) 	 & 6.2 (0.5) 	 & 54.7 (5.9) 	    & 4.68 (0.15) 	 & 40.6 (3.4) 	 & 31.2 (3.5) 	 & 11.7 (2.1) 	 \\
0.01	 &0.0	 &0.1	 & 2.26 (0.02) 	 & 8.5 (0.5) 	 & 6.0 (0.5) 	 & 43.4 (4.6) 	    & 4.76 (0.07) 	 & 37.2 (2.1) 	 & 27.5 (2.3) 	 & 7.5 (3.0) 	 \\
0.001	 &0.0	 &0.1	 & 2.26 (0.02) 	 & 8.4 (0.3) 	 & 6.0 (0.3) 	 & 48.0 (14.4) 	    & 4.71 (0.09) 	 & 40.9 (2.8) 	 & 31.7 (3.3) 	 & 10.9 (3.3) 	 \\
0.0	 &0.01	 &0.1	 & 2.26 (0.02) 	 & 8.0 (0.4) 	 & 5.5 (0.3) 	 & 53.1 (8.5) 	    & 4.74 (0.11) 	 & 40.4 (2.0) 	 & 31.2 (2.0) 	 & 7.7 (3.7) 	 \\
0.01	 &0.01	 &0.1	 & 2.27 (0.02) 	 & 8.1 (0.4) 	 & 5.7 (0.4) 	 & 49.1 (3.7) 	    & 4.76 (0.13) 	 & 36.0 (2.3) 	 & 26.0 (2.3) 	 & 6.4 (6.4) 	 \\
0.001	 &0.01	 &0.1	 & 2.24 (0.02) 	 & 8.1 (0.3) 	 & 5.7 (0.3) 	 & 49.8 (3.7) 	    & 4.76 (0.09) 	 & 40.7 (4.1) 	 & 31.8 (4.4) 	 & 10.9 (1.2) 	 \\
0.0	 &0.001	 &0.1	 & 2.25 (0.02) 	 & 8.3 (0.5) 	 & 6.0 (0.5) 	 & 45.2 (6.4) 	    & 4.75 (0.05) 	 & 39.5 (3.4) 	 & 30.5 (3.7) 	 & 9.6 (4.1) 	 \\
0.01	 &0.001	 &0.1	 & 2.26 (0.01) 	 & 8.2 (0.3) 	 & 5.8 (0.3) 	 & 58.2 (15.5) 	    & 4.81 (0.11) 	 & 37.2 (4.3) 	 & 27.5 (4.5) 	 & 10.4 (3.4) 	 \\
0.001	 &0.001	 &0.1	 & 2.25 (0.01) 	 & 8.3 (0.3) 	 & 6.0 (0.3) 	 & 43.8 (8.5) 	    & 4.80 (0.13) 	 & 38.8 (3.6) 	 & 29.5 (3.7) 	 & 11.6 (4.2) 	 \\
    \midrule
    \multicolumn{11}{c}{\wgan \quad $|B|=128$ \quad with \quad \mcd} \\ 
    \midrule
0.0	 &0.0	 &0.0	 & 2.25 (0.02) 	 & 7.7 (0.5) 	 & 5.4 (0.5) 	 & 20.2 (15.1) 	    & 4.75 (0.05) 	 & 38.6 (2.2) 	 & 28.7 (2.2) 	 & 18.7 (6.2) 	 \\   
0.01	 &0.0	 &0.0	 & 2.26 (0.02) 	 & 7.8 (0.3) 	 & 5.4 (0.3) 	 & 30.5 (11.8) 	    & 4.87 (0.20) 	 & 32.8 (2.9) 	 & 22.1 (2.8) 	 & 13.6 (8.0) 	 \\   
0.001	 &0.0	 &0.0	 & 2.24 (0.01) 	 & 8.0 (0.3) 	 & 5.6 (0.4) 	 & 34.1 (8.3) 	    & 4.79 (0.08) 	 & 40.2 (3.2) 	 & 29.5 (3.2) 	 & 14.1 (6.3) 	 \\   
0.0	 &0.01	 &0.0	 & 2.25 (0.01) 	 & 8.1 (0.8) 	 & 5.7 (0.8) 	 & 19.7 (8.7) 	    & 4.72 (0.07) 	 & 39.8 (3.7) 	 & 29.2 (3.9) 	 & 19.0 (8.1) 	 \\   
0.01	 &0.01	 &0.0	 & 2.26 (0.02) 	 & 7.8 (0.3) 	 & 5.4 (0.3) 	 & 28.3 (11.4) 	    & 4.77 (0.15) 	 & 35.9 (2.6) 	 & 25.1 (2.5) 	 & 23.8 (12.6) 	 \\   
0.001	 &0.01	 &0.0	 & 2.25 (0.02) 	 & 7.8 (0.2) 	 & 5.5 (0.3) 	 & 31.6 (10.0) 	    & 4.73 (0.08) 	 & 38.7 (1.9) 	 & 28.2 (1.8) 	 & 10.4 (6.4) 	 \\   
0.0	 &0.001	 &0.0	 & 2.24 (0.01) 	 & 7.7 (0.4) 	 & 5.5 (0.4) 	 & 31.7 (10.9) 	    & 4.80 (0.09) 	 & 37.7 (3.3) 	 & 27.0 (3.3) 	 & 15.2 (10.7) 	 \\   
0.01	 &0.001	 &0.0	 & 2.24 (0.01) 	 & 8.1 (0.4) 	 & 5.8 (0.4) 	 & 33.3 (7.2) 	    & 4.73 (0.08) 	 & 33.9 (2.7) 	 & 23.3 (3.0) 	 & 9.7 (4.2) 	 \\   
0.001	 &0.001	 &0.0	 & 2.25 (0.01) 	 & 8.0 (0.7) 	 & 5.7 (0.7) 	 & 32.5 (8.4) 	    & 4.66 (0.15) 	 & 40.6 (1.9) 	 & 29.9 (2.2) 	 & 22.7 (10.7) 	 \\   
0.0	 &0.0	 &0.01	 & 2.25 (0.02) 	 & 7.8 (0.4) 	 & 5.4 (0.4) 	 & 35.6 (5.7) 	    & 4.75 (0.12) 	 & 38.0 (3.6) 	 & 27.2 (3.3) 	 & 16.1 (5.3) 	 \\   
0.01	 &0.0	 &0.01	 & 2.26 (0.02) 	 & 7.9 (0.2) 	 & 5.5 (0.2) 	 & 27.8 (15.7) 	    & 4.81 (0.07) 	 & 34.5 (2.8) 	 & 23.7 (3.1) 	 & 17.9 (11.2) 	 \\   
0.001	 &0.0	 &0.01	 & 2.25 (0.02) 	 & 7.8 (0.5) 	 & 5.5 (0.6) 	 & 24.3 (9.9) 	    & 4.80 (0.08) 	 & 39.2 (4.4) 	 & 28.8 (4.1) 	 & 18.6 (6.9) 	 \\   
0.0	 &0.01	 &0.01	 & 2.25 (0.02) 	 & 7.6 (0.4) 	 & 5.2 (0.4) 	 & 32.8 (7.7) 	    & 4.77 (0.15) 	 & 39.3 (3.1) 	 & 28.6 (2.7) 	 & 15.8 (3.2) 	 \\   
0.01	 &0.01	 &0.01	 & 2.25 (0.01) 	 & 7.6 (0.3) 	 & 5.3 (0.2) 	 & 24.3 (7.3) 	    & 4.82 (0.09) 	 & 33.2 (1.4) 	 & 22.4 (1.4) 	 & 24.0 (8.8) 	 \\   
0.001	 &0.01	 &0.01	 & 2.25 (0.02) 	 & 8.0 (0.4) 	 & 5.8 (0.3) 	 & 30.0 (5.7) 	    & 4.69 (0.06) 	 & 37.3 (2.2) 	 & 26.8 (2.3) 	 & 14.3 (5.1) 	 \\   
0.0	 &0.001	 &0.01	 & 2.25 (0.02) 	 & 7.8 (0.3) 	 & 5.4 (0.3) 	 & 31.0 (12.6) 	    & 4.67 (0.09) 	 & 38.8 (3.7) 	 & 28.4 (3.8) 	 & 20.0 (5.1) 	 \\   
0.01	 &0.001	 &0.01	 & 2.23 (0.02) 	 & 7.8 (0.4) 	 & 5.5 (0.4) 	 & 25.3 (4.8) 	    & 4.84 (0.17) 	 & 34.4 (1.8) 	 & 23.6 (1.8) 	 & 19.4 (9.2) 	 \\   
0.001	 &0.001	 &0.01	 & 2.25 (0.01) 	 & 8.1 (0.3) 	 & 5.7 (0.3) 	 & 26.3 (6.4) 	    & 4.80 (0.05) 	 & 38.3 (2.9) 	 & 27.8 (3.2) 	 & 14.0 (2.2) 	 \\   
0.0	 &0.0	 &0.1	 & 2.25 (0.01) 	 & 7.6 (0.4) 	 & 5.2 (0.4) 	 & 24.2 (4.2) 	    & 4.86 (0.08) 	 & 38.9 (3.3) 	 & 28.4 (3.5) 	 & 14.4 (3.7) 	 \\   
0.01	 &0.0	 &0.1	 & 2.25 (0.02) 	 & 7.7 (0.3) 	 & 5.2 (0.3) 	 & 28.0 (11.9) 	    & 4.76 (0.10) 	 & 35.3 (3.3) 	 & 24.9 (3.4) 	 & 12.4 (6.8) 	 \\   
0.001	 &0.0	 &0.1	 & 2.25 (0.01) 	 & 7.8 (0.3) 	 & 5.4 (0.2) 	 & 17.4 (4.2) 	    & 4.74 (0.07) 	 & 40.6 (2.2) 	 & 30.1 (2.1) 	 & 12.4 (5.3) 	 \\   
0.0	 &0.01	 &0.1	 & 2.25 (0.02) 	 & 7.7 (0.2) 	 & 5.3 (0.2) 	 & 26.0 (2.2) 	    & 4.78 (0.11) 	 & 39.0 (3.9) 	 & 28.6 (3.8) 	 & 16.1 (8.4) 	 \\   
0.01	 &0.01	 &0.1	 & 2.25 (0.02) 	 & 7.2 (0.1) 	 & 4.8 (0.2) 	 & 20.2 (7.5) 	    & 4.76 (0.06) 	 & 36.0 (2.5) 	 & 25.3 (2.6) 	 & 11.8 (4.9) 	 \\   
0.001	 &0.01	 &0.1	 & 2.25 (0.01) 	 & 7.8 (0.3) 	 & 5.4 (0.3) 	 & 29.8 (4.9) 	    & 4.71 (0.10) 	 & 41.8 (2.6) 	 & 31.3 (2.6) 	 & 14.2 (7.7) 	 \\   
0.0	 &0.001	 &0.1	 & 2.24 (0.01) 	 & 7.5 (0.3) 	 & 5.1 (0.3) 	 & 21.6 (14.4) 	    & 4.77 (0.12) 	 & 38.9 (2.3) 	 & 28.4 (2.4) 	 & 14.4 (4.0) 	 \\   
0.01	 &0.001	 &0.1	 & 2.25 (0.02) 	 & 7.9 (0.3) 	 & 5.5 (0.3) 	 & 21.8 (11.5) 	    & 4.72 (0.07) 	 & 37.9 (2.8) 	 & 27.0 (2.6) 	 & 7.0 (3.7) 	 \\   
0.001	 &0.001	 &0.1	 & 2.25 (0.02) 	 & 7.7 (0.3) 	 & 5.4 (0.3) 	 & 28.7 (6.8) 	    & 4.72 (0.07) 	 & 40.6 (5.4) 	 & 30.7 (5.5) 	 & 13.4 (3.5) 	 \\   
    \bottomrule
  \end{tabular}}
  \end{sc}
  \end{small}
  \end{center}
\end{table}

\begin{table}[!h]
  \caption{\dcgan architecture with Wasserstein divergence objective \citep{Wu18} on \ffhqlowres and \celeba. Standard deviations, calculated over five seeds times three repetitions of sampling $10\,000$ images, are reported in parenthesis.
  }
  \label{tab:res:all_ffhq128_celeba}
  \begin{center}
    \begin{small}
      \begin{sc}
\resizebox{\textwidth}{!}{\begin{tabular}{lll|llll|llll}
    \toprule
    & & & \multicolumn{4}{c|}{\ffhqlowres} &  \multicolumn{4}{c}{\celeba} \\
    \midrule    
    \multicolumn{11}{c}{\wgan \quad $|B|=128$} \\ 
    \midrule
  $\sigma^2_{\mathrm{lik}}$  &  $\lambda_{\mathrm{gr}}$  & $\rho_{\mathrm{SAM}}$  &   
    \iscscore $\uparrow$   &  \fidscore $\downarrow$  &  \kidscore {\scriptsize $\, \times 10^{-3} \,$} $\downarrow$  &
    $\ell_1$ & 
    \iscscore $\uparrow$   &  \fidscore $\downarrow$  &  \kidscore {\scriptsize $\, \times 10^{-3} \,$} $\downarrow$  &
    $\ell_1$ \\ 
    \midrule
0.0	 &0.0	 &0.0	 & 3.21 (0.05) 	 & 63.6 (1.5) 	 & 55.8 (1.5) 	 & 45.9 (13.7) 	    & 2.87 (0.04) 	 & 18.3 (0.9) 	 & 12.4 (1.1) 	 & 31.2 (9.9) 	 \\
0.01	 &0.0	 &0.0	 & 3.25 (0.04) 	 & 62.8 (1.5) 	 & 54.8 (1.6) 	 & 51.6 (19.4) 	    & 2.89 (0.05) 	 & 17.8 (2.6) 	 & 11.5 (2.7) 	 & 26.4 (11.7) 	 \\
0.001	 &0.0	 &0.0	 & 3.24 (0.03) 	 & 63.2 (0.9) 	 & 55.6 (0.7) 	 & 49.7 (13.5) 	    & 2.87 (0.03) 	 & 18.4 (0.6) 	 & 12.4 (0.6) 	 & 25.6 (13.3) 	 \\
0.0	 &0.01	 &0.0	 & 3.22 (0.04) 	 & 65.1 (3.5) 	 & 57.0 (3.4) 	 & 37.9 (9.4) 	    & 2.88 (0.03) 	 & 18.5 (0.9) 	 & 12.6 (1.1) 	 & 33.4 (8.8) 	 \\
0.01	 &0.01	 &0.0	 & 3.24 (0.03) 	 & 63.5 (2.6) 	 & 55.7 (2.8) 	 & 48.5 (20.9) 	    & 2.87 (0.08) 	 & 24.8 (16.9) 	 & 18.6 (16.6) 	 & 30.0 (15.8) 	 \\
0.001	 &0.01	 &0.0	 & 3.20 (0.04) 	 & 66.3 (3.0) 	 & 58.3 (2.8) 	 & 43.2 (17.2) 	    & 2.88 (0.03) 	 & 18.1 (0.8) 	 & 12.0 (0.9) 	 & 22.9 (11.5) 	 \\
0.0	 &0.001	 &0.0	 & 3.24 (0.05) 	 & 62.9 (1.9) 	 & 54.9 (1.9) 	 & 43.4 (14.6) 	    & 2.88 (0.02) 	 & 17.9 (0.8) 	 & 11.9 (0.8) 	 & 18.0 (11.0) 	 \\
0.01	 &0.001	 &0.0	 & 3.26 (0.07) 	 & 62.1 (3.0) 	 & 54.2 (3.5) 	 & 33.8 (10.2) 	    & 2.90 (0.03) 	 & 25.4 (19.1) 	 & 19.0 (18.4) 	 & 35.9 (13.3) 	 \\
0.001	 &0.001	 &0.0	 & 3.22 (0.04) 	 & 64.9 (2.2) 	 & 57.5 (2.7) 	 & 31.5 (24.9) 	    & 2.88 (0.03) 	 & 17.2 (0.8) 	 & 11.1 (0.9) 	 & 32.4 (12.0) 	 \\
0.0	 &0.0	 &0.01	 & 3.22 (0.04) 	 & 62.9 (3.5) 	 & 55.4 (4.0) 	 & 38.2 (14.3) 	    & 2.89 (0.05) 	 & 28.2 (18.0) 	 & 21.7 (17.2) 	 & 20.2 (10.5) 	 \\
0.01	 &0.0	 &0.01	 & 3.25 (0.03) 	 & 61.6 (1.4) 	 & 53.8 (1.5) 	 & 46.0 (12.5) 	    & 2.90 (0.05) 	 & 18.4 (3.7) 	 & 12.2 (3.4) 	 & 26.6 (14.4) 	 \\
0.001	 &0.0	 &0.01	 & 3.21 (0.04) 	 & 64.6 (4.7) 	 & 56.7 (4.9) 	 & 29.9 (17.1) 	    & 2.88 (0.03) 	 & 17.8 (0.7) 	 & 12.1 (0.6) 	 & 39.9 (5.5) 	 \\
0.0	 &0.01	 &0.01	 & 3.22 (0.06) 	 & 63.1 (3.9) 	 & 55.2 (4.0) 	 & 54.6 (16.8) 	    & 2.91 (0.02) 	 & 18.4 (1.1) 	 & 12.5 (1.2) 	 & 32.8 (8.2) 	 \\
0.01	 &0.01	 &0.01	 & 3.26 (0.03) 	 & 61.2 (2.5) 	 & 53.0 (2.6) 	 & 40.1 (16.9) 	    & 2.90 (0.03) 	 & 15.9 (0.5) 	 & 10.0 (0.6) 	 & 38.0 (13.2) 	 \\
0.001	 &0.01	 &0.01	 & 3.22 (0.03) 	 & 62.4 (3.1) 	 & 54.4 (3.5) 	 & 31.2 (19.9) 	    & 2.90 (0.05) 	 & 17.4 (0.9) 	 & 11.5 (1.0) 	 & 31.3 (14.1) 	 \\
0.0	 &0.001	 &0.01	 & 3.24 (0.03) 	 & 61.8 (2.6) 	 & 53.8 (2.8) 	 & 43.9 (8.5) 	    & 2.91 (0.04) 	 & 17.7 (0.6) 	 & 11.7 (0.7) 	 & 29.5 (11.6) 	 \\
0.01	 &0.001	 &0.01	 & 3.24 (0.03) 	 & 64.3 (2.1) 	 & 56.7 (2.4) 	 & 33.7 (18.7) 	    & 2.91 (0.03) 	 & 16.4 (1.0) 	 & 10.5 (1.2) 	 & 43.6 (10.4) 	 \\
0.001	 &0.001	 &0.01	 & 3.23 (0.03) 	 & 62.5 (1.5) 	 & 54.8 (1.8) 	 & 50.9 (17.9) 	    & 2.89 (0.02) 	 & 18.2 (1.0) 	 & 12.3 (1.0) 	 & 33.2 (11.6) 	 \\
0.0	 &0.0	 &0.1	 & 3.26 (0.05) 	 & 62.1 (2.0) 	 & 53.9 (2.3) 	 & 37.1 (10.9) 	    & 2.84 (0.09) 	 & 34.9 (20.8) 	 & 28.5 (20.1) 	 & 19.3 (17.4) 	 \\
0.01	 &0.0	 &0.1	 & 3.27 (0.03) 	 & 60.8 (2.7) 	 & 52.8 (2.8) 	 & 47.2 (9.4) 	    & 2.86 (0.12) 	 & 29.7 (16.7) 	 & 23.3 (16.0) 	 & 18.3 (11.4) 	 \\
0.001	 &0.0	 &0.1	 & 3.25 (0.04) 	 & 60.9 (2.3) 	 & 53.1 (2.9) 	 & 43.4 (12.5) 	    & 2.85 (0.09) 	 & 25.8 (16.1) 	 & 20.1 (15.4) 	 & 26.1 (7.0) 	 \\
0.0	 &0.01	 &0.1	 & 3.22 (0.04) 	 & 62.1 (2.4) 	 & 54.2 (2.5) 	 & 30.6 (17.0) 	    & 2.90 (0.08) 	 & 37.6 (24.0) 	 & 32.6 (25.0) 	 & 18.6 (6.3) 	 \\
0.01	 &0.01	 &0.1	 & 3.27 (0.04) 	 & 58.5 (1.0) 	 & 50.1 (1.4) 	 & 40.4 (13.8) 	    & 2.86 (0.09) 	 & 29.0 (17.3) 	 & 22.1 (16.0) 	 & 28.5 (18.6) 	 \\
0.001	 &0.01	 &0.1	 & 3.21 (0.05) 	 & 63.0 (2.5) 	 & 55.2 (2.7) 	 & 47.6 (14.5) 	    & 2.85 (0.07) 	 & 35.3 (20.8) 	 & 29.7 (20.5) 	 & 25.6 (14.2) 	 \\
0.0	 &0.001	 &0.1	 & 3.20 (0.04) 	 & 62.9 (4.7) 	 & 55.2 (4.6) 	 & 40.0 (11.1) 	    & 2.87 (0.08) 	 & 24.6 (13.2) 	 & 18.6 (12.2) 	 & 27.9 (17.8) 	 \\
0.01	 &0.001	 &0.1	 & 3.26 (0.03) 	 & 60.2 (1.9) 	 & 51.8 (2.1) 	 & 34.4 (8.7) 	    & 2.87 (0.13) 	 & 29.5 (16.8) 	 & 23.4 (16.7) 	 & 21.1 (16.2) 	 \\
0.001	 &0.001	 &0.1	 & 3.25 (0.05) 	 & 63.2 (2.5) 	 & 55.0 (2.5) 	 & 38.8 (10.6) 	    & 2.87 (0.06) 	 & 31.4 (16.0) 	 & 24.8 (14.7) 	 & 10.0 (3.2) 	 \\
    \midrule
    \multicolumn{11}{c}{\wgan \quad $|B|=128$ \quad with \quad \mcd} \\ 
    \midrule
0.0	 &0.0	 &0.0	 & 3.20 (0.06) 	 & 65.5 (3.6) 	 & 57.4 (3.6) 	 & 36.1 (19.0) 	     & 2.95 (0.13) 	 & 42.9 (19.4) 	 & 34.9 (18.2) 	 & 24.1 (8.8) 	 \\
0.01	 &0.0	 &0.0	 & 3.24 (0.04) 	 & 67.5 (3.0) 	 & 59.1 (2.5) 	 & 46.3 (8.3) 	     & 2.83 (0.12) 	 & 60.8 (22.0) 	 & 55.3 (22.6) 	 & 19.5 (9.0) 	 \\
0.001	 &0.0	 &0.0	 & 3.19 (0.04) 	 & 63.1 (3.4) 	 & 55.1 (3.9) 	 & 48.4 (14.0) 	     & 3.01 (0.11) 	 & 65.0 (23.1) 	 & 58.3 (22.8) 	 & 20.0 (17.1) 	 \\
0.0	 &0.01	 &0.0	 & 3.17 (0.04) 	 & 67.4 (3.7) 	 & 59.3 (3.4) 	 & 49.7 (19.1) 	     & 3.02 (0.11) 	 & 81.1 (8.1) 	 & 75.7 (11.2) 	 & 25.1 (17.7) 	 \\
0.01	 &0.01	 &0.0	 & 3.25 (0.05) 	 & 62.8 (1.5) 	 & 55.1 (2.0) 	 & 44.2 (15.4) 	     & 2.83 (0.14) 	 & 70.4 (8.8) 	 & 64.1 (9.9) 	 & 29.0 (29.6) 	 \\
0.001	 &0.01	 &0.0	 & 3.20 (0.06) 	 & 65.0 (1.8) 	 & 56.9 (1.5) 	 & 40.3 (18.2) 	     & 2.95 (0.17) 	 & 76.1 (2.7) 	 & 70.3 (4.9) 	 & 33.4 (11.6) 	 \\
0.0	 &0.001	 &0.0	 & 3.19 (0.03) 	 & 66.7 (3.0) 	 & 58.4 (3.6) 	 & 20.3 (16.7) 	     & 2.96 (0.08) 	 & 59.8 (27.6) 	 & 53.8 (28.8) 	 & 31.0 (11.3) 	 \\
0.01	 &0.001	 &0.0	 & 3.23 (0.04) 	 & 64.6 (3.5) 	 & 56.8 (3.4) 	 & 48.0 (7.8) 	     & 2.93 (0.08) 	 & 63.5 (24.5) 	 & 57.4 (25.8) 	 & 19.7 (9.0) 	 \\
0.001	 &0.001	 &0.0	 & 3.21 (0.04) 	 & 64.3 (2.1) 	 & 56.1 (2.3) 	 & 27.0 (11.3) 	     & 3.11 (0.06) 	 & 76.9 (6.1) 	 & 67.8 (7.1) 	 & 28.7 (6.8) 	 \\
0.0	 &0.0	 &0.01	 & 3.19 (0.05) 	 & 68.9 (2.5) 	 & 60.8 (2.7) 	 & 46.1 (13.3) 	     & 2.98 (0.11) 	 & 68.3 (25.6) 	 & 63.0 (26.4) 	 & 27.8 (5.3) 	 \\
0.01	 &0.0	 &0.01	 & 3.22 (0.06) 	 & 66.6 (2.7) 	 & 58.1 (3.2) 	 & 35.6 (21.2) 	     & 2.94 (0.07) 	 & 66.0 (11.5) 	 & 57.1 (15.3) 	 & 32.5 (22.4) 	 \\
0.001	 &0.0	 &0.01	 & 3.20 (0.04) 	 & 67.7 (1.6) 	 & 59.7 (1.8) 	 & 57.6 (11.4) 	     & 2.88 (0.07) 	 & 41.7 (24.5) 	 & 34.8 (24.6) 	 & 14.6 (8.0) 	 \\
0.0	 &0.01	 &0.01	 & 3.19 (0.07) 	 & 65.3 (4.2) 	 & 57.2 (4.4) 	 & 38.3 (13.1) 	     & 2.92 (0.15) 	 & 75.3 (14.6) 	 & 71.0 (17.3) 	 & 29.5 (17.3) 	 \\
0.01	 &0.01	 &0.01	 & 3.21 (0.05) 	 & 65.3 (3.7) 	 & 57.5 (3.8) 	 & 59.8 (27.0) 	     & 2.84 (0.06) 	 & 44.8 (25.5) 	 & 37.9 (26.1) 	 & 20.3 (16.4) 	 \\
0.001	 &0.01	 &0.01	 & 3.19 (0.04) 	 & 63.3 (2.0) 	 & 55.0 (1.9) 	 & 77.7 (28.9) 	     & 2.88 (0.06) 	 & 81.8 (8.4) 	 & 80.1 (10.9) 	 & 14.0 (5.3) 	 \\
0.0	 &0.001	 &0.01	 & 3.18 (0.03) 	 & 63.2 (3.0) 	 & 55.2 (3.4) 	 & 61.2 (29.4) 	     & 2.81 (0.12) 	 & 69.7 (5.2) 	 & 64.2 (6.2) 	 & 20.7 (5.2) 	 \\
0.01	 &0.001	 &0.01	 & 3.23 (0.06) 	 & 67.2 (4.8) 	 & 59.0 (4.5) 	 & 46.6 (22.0) 	     & 2.89 (0.06) 	 & 52.5 (29.1) 	 & 47.4 (30.2) 	 & 23.5 (9.1) 	 \\
0.001	 &0.001	 &0.01	 & 3.21 (0.09) 	 & 66.8 (5.0) 	 & 58.3 (5.3) 	 & 86.0 (95.1) 	     & 2.80 (0.14) 	 & 59.9 (21.7) 	 & 54.2 (21.5) 	 & 26.3 (10.0) 	 \\
0.0	 &0.0	 &0.1	 & 3.19 (0.04) 	 & 66.5 (4.1) 	 & 58.1 (4.2) 	 & 42.0 (19.0) 	     & 2.89 (0.08) 	 & 40.1 (27.0) 	 & 34.2 (27.5) 	 & 29.2 (6.9) 	 \\
0.01	 &0.0	 &0.1	 & 3.23 (0.04) 	 & 62.8 (3.1) 	 & 54.7 (3.5) 	 & 50.8 (16.4) 	     & 2.83 (0.11) 	 & 51.5 (29.3) 	 & 47.7 (31.6) 	 & 14.7 (8.0) 	 \\
0.001	 &0.0	 &0.1	 & 3.18 (0.04) 	 & 64.5 (1.8) 	 & 56.1 (1.8) 	 & 38.3 (26.5) 	     & 2.98 (0.12) 	 & 60.1 (23.7) 	 & 52.7 (24.0) 	 & 17.4 (10.9) 	 \\
0.0	 &0.01	 &0.1	 & 3.21 (0.05) 	 & 63.7 (4.8) 	 & 55.3 (5.1) 	 & 27.7 (18.1) 	     & 2.81 (0.10) 	 & 49.8 (25.7) 	 & 45.2 (26.7) 	 & 24.1 (15.3) 	 \\
0.01	 &0.01	 &0.1	 & 3.23 (0.04) 	 & 62.6 (4.0) 	 & 54.7 (4.2) 	 & 46.2 (24.3) 	     & 2.95 (0.06) 	 & 67.0 (25.9) 	 & 60.8 (26.5) 	 & 24.1 (15.0) 	 \\
0.001	 &0.01	 &0.1	 & 3.20 (0.03) 	 & 61.7 (1.2) 	 & 52.7 (1.0) 	 & 43.1 (23.3) 	     & 2.87 (0.13) 	 & 68.2 (16.7) 	 & 63.8 (19.6) 	 & 12.6 (6.8) 	 \\
0.0	 &0.001	 &0.1	 & 3.21 (0.06) 	 & 61.7 (2.2) 	 & 53.1 (2.3) 	 & 55.3 (20.8) 	     & 2.85 (0.13) 	 & 49.4 (27.8) 	 & 43.4 (28.7) 	 & 28.8 (24.9) 	 \\
0.01	 &0.001	 &0.1	 & 3.23 (0.04) 	 & 61.4 (2.6) 	 & 53.0 (2.9) 	 & 36.0 (19.2) 	     & 2.84 (0.09) 	 & 50.0 (27.1) 	 & 43.3 (26.9) 	 & 17.8 (16.9) 	 \\
0.001	 &0.001	 &0.1	 & 3.19 (0.06) 	 & 65.6 (5.4) 	 & 56.7 (5.5) 	 & 61.4 (20.4) 	     & 2.94 (0.18) 	 & 74.5 (7.8) 	 & 69.0 (10.8) 	 & 13.1 (6.5) 	 \\
   \bottomrule
  \end{tabular}}
  \end{sc}
  \end{small}
  \end{center}
\end{table}

\newpage

\

\newpage

\section{Results on \rgans}

In this section, we report results on \rgans, which we generally found to be more unstable to train compared to $\wgans$.

\begin{table}[H]
  \caption{\dcgan architecture with the objective of \rgans \cite{jolicoeur-martineau2018} on \mnist, \cifar, \ffhqlowres, and \celeba. Standard deviations, calculated over five seeds times three repetitions of sampling $10\,000$ images, are reported in parenthesis.
  The arrows indicate whether metrics are so that the higher the better ($\uparrow$) or the lower the better ($\downarrow$).
  $|B|$ denotes the batch size. 
  }
  \label{tab:res:summary:rgans}
  \begin{center}
    \begin{small}
      \begin{sc}
\resizebox{\textwidth}{!}{\begin{tabular}{l|llll|llll}
    \toprule
    \multicolumn{9}{c}{\rgan \quad $|B|=128$} \\ 
    \midrule
     & \multicolumn{4}{c|}{\mnist} &  \multicolumn{4}{c}{\cifar} \\
    \midrule    
  Strategy  &   
    \iscscore $\uparrow$   &  \fidscore $\downarrow$  &  \kidscore {\scriptsize $\, \times 10^{-3} \,$} $\downarrow$  &
    $\ell_1$ & 
    \iscscore $\uparrow$   &  \fidscore $\downarrow$  &  \kidscore {\scriptsize $\, \times 10^{-3} \,$} $\downarrow$  &
    $\ell_1$ \\ 
    \midrule
None	 & 1.93 (0.46) 	 & 101.8 (184.3) 	 & 134.7 (257.7) 	 & 1165.7 (1313.3) 	 	 & 5.01 (0.12) 	 & 21.0 (1.7)   	 & 10.8 (1.5)   	 & 63.4 (30.8)  	 \\      
Likelihood relax. ($\sigma^2_{\mathrm{lik}} = 0.01$) 	 & 1.79 (0.67) 	 & 251.8 (172.3) 	 & 306.2 (272.1) 	 & 1582.6 (838.7) 	  & 2.40 (0.40) 	 & 239.3 (33.0) 	 & 168.7 (37.9) 	 & 6692.6 (6632.2) 	 \\      
Gradient reg.	($\lambda_{\mathrm{gr}} = 0.01$)	 & 1.92 (0.46) 	 & 111.5 (141.1) 	 & 119.9 (191.0) 	 & 2774.3 (2475.6) 	 	 & 5.16 (0.10) 	 & 19.6 (1.4)   	 & 9.4 (0.8)    	 & 61.1 (34.2) 	         \\      
\sam ($\rho_{\mathrm{SAM}} = 0.1$)	 & 2.14 (0.02) 	 & 9.2 (0.9)    	 & 5.7 (0.8)    	 & 2802.8 (3767.8) 	 	 & 5.19 (0.13) 	 & 19.7 (0.7)   	 & 9.9 (0.5)    	 & 65.0 (42.1)  	 \\      
\mcd ($p = 0.1$) 	 & 2.11 (0.11) 	 & 50.3 (49.6)  	 & 41.9 (46.4)  	 & 1330.1 (1174.8) 	  & 4.92 (0.05) 	 & 21.6 (0.2)   	 & 11.8 (0.3)   	 & 918.8 (588.8) 	 \\   
    \midrule
    \midrule
     & \multicolumn{4}{c|}{\ffhqlowres} &  \multicolumn{4}{c}{\celeba} \\
    \midrule    
Strategy  &   
    \iscscore $\uparrow$   &  \fidscore $\downarrow$  &  \kidscore {\scriptsize $\, \times 10^{-3} \,$} $\downarrow$  &
    $\ell_1$ & 
    \iscscore $\uparrow$   &  \fidscore $\downarrow$  &  \kidscore {\scriptsize $\, \times 10^{-3} \,$} $\downarrow$  &
    $\ell_1$ \\ 
    \midrule
None      & 3.38 (0.05) 	 & 54.2 (4.8) 	 & 42.9 (4.5) 	 & 65.8 (50.7) 	     & 2.82 (0.10) 	 & 22.6 (2.2) 	 & 9.7 (0.8) 	 & 37.3 (19.8) 	   \\
Likelihood relax. ($\sigma^2_{\mathrm{lik}} = 0.01$) 	 & 3.08 (0.25) 	 & 91.2 (24.0) 	 & 73.6 (18.8) 	 & 154.8 (203.3)     & 1.32 (0.32) 	 & 399.3 (44.8)  & 471.0 (79.9)  & 3909.8 (4234.7) \\
Gradient reg.	($\lambda_{\mathrm{gr}} = 0.01$)	 & 3.34 (0.10) 	 & 56.2 (5.2) 	 & 45.3 (6.6) 	 & 34.7 (28.2) 	     & 2.79 (0.08) 	 & 23.0 (2.2) 	 & 9.5 (1.4) 	 & 66.7 (47.6) 	   \\
\sam ($\rho_{\mathrm{SAM}} = 0.1$)	 & 3.33 (0.10) 	 & 52.7 (3.0) 	 & 41.1 (4.3) 	 & 50.3 (17.2) 	     & 2.77 (0.03) 	 & 24.3 (2.6) 	 & 10.4 (2.0) 	 & 37.6 (38.0) 	   \\
\mcd ($p = 0.1$) 	 & 3.34 (0.11) 	 & 53.0 (2.7) 	 & 44.7 (2.8) 	 & 80.4 (86.7) 	     & 2.76 (0.03) 	 & 20.1 (1.8) 	 & 11.1 (1.9) 	 & 59.9 (49.9) 	   \\

   \bottomrule
  \end{tabular}}
  \end{sc}
  \end{small}
  \end{center}
\end{table}


\end{document}